\documentclass{article}

\usepackage{arxiv}

\usepackage[utf8]{inputenc} 
\usepackage[T1]{fontenc}    
\usepackage{hyperref}       
\usepackage{url}            
\usepackage{booktabs}       
\usepackage{amsfonts}       
\usepackage{nicefrac}       
\usepackage{microtype}      

\usepackage{lipsum}         
\usepackage{graphicx}
\usepackage{natbib}
\usepackage{doi}

\usepackage{authblk}

\usepackage{graphicx}%
\usepackage{multirow}%
\usepackage{amsmath,amssymb,amsfonts}%
\usepackage{amsthm}%
\usepackage{mathrsfs}%
\usepackage[title]{appendix}%
\usepackage{xcolor}%
\usepackage{textcomp}%
\usepackage{manyfoot}%
\usepackage{booktabs}%
\usepackage{algorithm}%
\usepackage{algorithmicx}%
\usepackage{algpseudocode}%
\usepackage{listings}%
\usepackage{indentfirst}
\usepackage{placeins}
\usepackage{booktabs}
\usepackage{diagbox} %
\usepackage{caption} %
\usepackage{hyperref}
\usepackage{float}
\newtheorem{property}{Property}
\usepackage{cleveref}       

\title{Deep learning models are vulnerable, but adversarial examples are even more vulnerable}

\newif\ifuniqueAffiliation
\uniqueAffiliationtrue

\ifuniqueAffiliation 

\ifdefined\acl
\author{ 
	\href{https://orcid.org/0000-0003-3387-695X}{\includegraphics[scale=0.06]{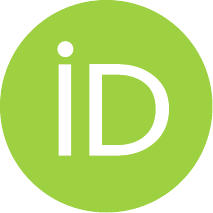}\hspace{1mm}Jun Li}\textsuperscript{$\dagger$}\thanks{Corresponding author: \texttt{lijun@jlufe.edu.cn}} \\
	School of Management Science and Information Engineering\\
	Jilin University of Finance and Economics\\
	Jingyue Street, Changchun 130117, China \\
	\texttt{lijun@jlufe.edu.cn} \\
	\And
	\href{https://orcid.org/0009-0009-0724-1900}{\includegraphics[scale=0.06]{orcid.pdf}\hspace{1mm}Yanwei Xu}\textsuperscript{*,$\dagger$} \\
	School of Management Science and Information Engineering\\
	Jilin University of Finance and Economics\\
	Jingyue Street, Changchun 130117, China \\
	\texttt{xuyanwei@s.jlufe.edu.cn} \\
	\And
	\href{https://orcid.org/0009-0004-6571-5119}{\includegraphics[scale=0.06]{orcid.pdf}\hspace{1mm}Keran Li} \\
	School of Management Science and Information Engineering\\
	Jilin University of Finance and Economics\\
	Jingyue Street, Changchun 130117, China \\
	\texttt{13624158722@163.com} \\
	\And
	\href{https://orcid.org/0000-0001-8412-4956}{\includegraphics[scale=0.06]{orcid.pdf}\hspace{1mm}Xiaoli Zhang} \\
	College of Computer Science and Technology\\
	Jilin University\\
	Qianjin Street, Changchun 130012, China \\
	\texttt{zhangxiaoli@jlu.edu.cn} \\
}
\else
\usepackage{authblk}

\newbox{\orcid}\sbox{\orcid}{\includegraphics[scale=0.06]{orcid.pdf}} 
\author[1,2]{%
	\href{https://orcid.org/0000-0003-3387-695X}{\usebox{\orcid}\hspace{1mm}Jun Li}\textsuperscript{$\dagger$}%
}
\author[1]{%
	\href{https://orcid.org/0009-0009-0724-1900}{\usebox{\orcid}\hspace{1mm}Yanwei Xu}\textsuperscript{*,$\dagger$}%
}
\author[1]{%
	\href{https://orcid.org/0009-0004-6571-5119}{\usebox{\orcid}\hspace{1mm}Keran Li}%
}
\author[3]{%
	\href{https://orcid.org/0000-0001-8412-4956}{\usebox{\orcid}\hspace{1mm}Xiaoli Zhang}%
}

\affil[1]{School of Management Science and Information Engineering, Jilin University of Finance and Economics, Jingyue Street, Changchun 130117, China}
\affil[2]{Center for Artificial Intelligence, Jilin University of Finance and Economics, Jingyue Street, Changchun 130117, China}
\affil[3]{College of Computer Science and Technology, Jilin University, Qianjin Street, Changchun 130012, China}
\fi

\renewcommand{\shorttitle}{\textit{arXiv} Template}

\hypersetup{
pdftitle={A template for the arxiv style},
pdfsubject={q-bio.NC, q-bio.QM},
pdfauthor={David S.~Hippocampus, Elias D.~Striatum},
pdfkeywords={First keyword, Second keyword, More},
}

\begin{document}
\maketitle

\begin{abstract}
	\protect{Understanding intrinsic differences between adversarial examples and clean samples is key to enhancing DNN robustness and detection against adversarial attacks. This study first empirically finds that image-based adversarial examples are notably sensitive to occlusion. Controlled experiments on CIFAR-10 used nine canonical attacks (e.g., FGSM, PGD) to generate adversarial examples, paired with original samples for evaluation. We introduce Sliding Mask Confidence Entropy (SMCE) to quantify model confidence fluctuation under occlusion. Using 1800+ test images, SMCE calculations—supported by Mask Entropy Field Maps and statistical distributions—show adversarial examples have significantly higher confidence volatility under occlusion than originals. Based on this, we propose Sliding Window Mask-based Adversarial Example Detection (SWM-AED), which avoids catastrophic overfitting of conventional adversarial training. Evaluations across classifiers and attacks on CIFAR-10 demonstrate robust performance, with accuracy over 62\% in most cases and up to 96.5\%.}
	
\end{abstract}

\renewcommand{\thefootnote}{\fnsymbol{footnote}}
\footnotetext[1]{Corresponding author: \texttt{xuyanwei@s.jlufe.edu.cn}}
\footnotetext[2]{These authors contributed equally to this work.}


\section{Introduction}\label{sec1}

In the era of rapid digital transformation, artificial intelligence (AI), particularly deep learning, has become deeply integrated into diverse domains, including image recognition, speech processing, and natural language understanding, thereby significantly enhancing daily life and work. Deep learning models, renowned for their exceptional feature extraction and learning capabilities, have demonstrated outstanding performance in complex tasks. Critical applications, such as facial recognition in security systems, medical image diagnostics, and road condition monitoring in autonomous vehicles, heavily depend on the accurate decision-making of these models. However, as AI technology evolves, emerging security concerns surrounding deep learning systems have gained attention \citep{chakraborty2021survey}. \citep{szegedy2013intriguing} first introduced the concept of adversarial examples, wherein subtle, nearly imperceptible perturbations are added to input data, leading to drastic mispredictions by deep neural networks. This phenomenon underscores a fundamental divergence in how deep learning models and human cognition interpret data, while also posing significant threats to systems relying on AI. For instance, in the context of autonomous driving, adversarial attacks could mislead a vehicle into misidentifying traffic signs, potentially resulting in catastrophic accidents.

In response, researchers have developed a range of adversarial example generation techniques, including gradient-based attacks, optimization-based methods, and meta-learning-based approaches, etc. Notable methods include the Fast Gradient Sign Method (FGSM), Projected Gradient Descent (PGD) \citep{mkadry2017towards}, DeepFool \citep{r4}, and You Only Attack Once (YOAO) \citep{app15010302}, among others. These adversarial attacks are not confined to image classification tasks but have been extended to domains such as object detection, speech recognition, and real-world scenarios. To counteract the threats posed by adversarial examples, researchers have proposed various defense and detection strategies. Early defense mechanisms focused on adversarial training, where adversarial examples were incorporated into the model's training process. However, studies have shown that this approach often faces challenges, such as reduced detection accuracy and the risk of catastrophic overfitting during training. As research advanced, more sophisticated defense strategies emerged, including model ensembles that combine predictions from multiple models to enhance robustness. However, most defense methods still center on improving model robustness, which typically requires significant computational resources and increases model complexity.

To tackle the growing challenges of adversarial attacks, increasing attention has been directed toward adversarial example detection technologies. Unlike traditional defense methods, which focus on enhancing model robustness, adversarial example detection targets the identification and rejection of adversarial inputs before they can compromise the model, thereby bolstering the security of AI systems. These detection approaches typically offer lower computational complexity and reduced resource demands compared to robust training strategies, allowing for better cost control while effectively mitigating the impact of adversarial examples. The capacity to detect and filter adversarial examples prior to their integration holds considerable practical value, ensuring the reliability of deep learning models in real-world applications. Thus, establishing an efficient and accurate detection mechanism is paramount to preserving the integrity of AI-driven systems.

Despite the significant advancements in adversarial example detection, numerous challenges remain. As adversarial example generation techniques continuously evolve, attackers develop increasingly sophisticated perturbation strategies, which raise the difficulty of detection. New attack algorithms are adept at bypassing existing detection methods and exhibit greater concealment. Moreover, deep learning models themselves are highly complex and heterogeneous, with each model responding differently to adversarial inputs, complicating the development of universal detection models. To address these issues, this study delves into the causes of adversarial examples and the inherent vulnerabilities within deep learning models. While adversarial examples are often attributed to the weaknesses of these models, a more pressing issue lies in their inherent instability, whereby slight changes can lead to misclassifications across different categories. This phenomenon, termed the ``vulnerable of adversarial examples" represents a critical aspect that requires further investigation.

Building upon this finding, the study introduces the concept of Sliding Mask Confidence Entropy (SMCE) to quantify the vulnerable of adversarial examples and the stability of deep learning models. By applying a sliding window to mask local areas of an image and calculating the average of confidence entropy values for each window, the method assesses not only the stability of the image itself but also the robustness of the classifier when confronted with partial occlusion or perturbation. Through extensive experimentation, the study reveals a striking finding: as model detection accuracy improves, the use of the SMCE strategy results in enhanced robustness of the deep learning model, significantly boosting security. This approach effectively avoids the catastrophic overfitting associated with traditional adversarial training methods, thereby providing a novel research avenue for subsequent work in this domain.

Incorporating SMCE, the study proposes the Sliding Window Masking-Adversarial Example Detection (SWM-AED) algorithm, an innovative solution designed to improve model robustness for adversarial example detection. The core of this algorithm lies in leveraging SMCE values to detect adversarial examples. Specifically, the SWM-AED algorithm calculates the Sliding Mask Confidence Entropy of an image, identifying samples with abnormal entropy values indicative of adversarial perturbations. Notably, the SWM-AED algorithm is non-specific to any particular type of adversarial example, making it resilient to a broad range of adversarial attacks generated by various adversarial generation algorithms, demonstrating superior performance. The study further highlights a key observation: as the model's detection accuracy improves, the SWM-AED strategy adeptly mitigates catastrophic overfitting, enhancing both robustness and security. From a theoretical perspective, this study establishes a positive correlation between the accuracy and robustness of deep learning models after the integration of the SWM-AED algorithm.

This work fills a critical gap in the measurement of adversarial example vulnerable by introducing the concept of SMCE. Based on this metric, the SWM-AED detection algorithm demonstrates outstanding detection performance, achieving over 80\% accuracy in detecting adversarial examples generated by multiple attack algorithms. Moreover, by proactively identifying and filtering adversarial examples, the SWM-AED algorithm significantly mitigates their impact on deep learning models, thereby substantially improving the overall security of AI systems.

\section{Background and Related Work}\label{sec2}

Adversarial examples are generated by introducing subtle perturbations to an original image, resulting in misclassifications by deep neural networks (DNNs). Despite these perturbations being imperceptible to human observers, they cause the model to misidentify the original input—incorrectly assigning it a label B instead of the true label A. The vulnerability is particularly concerning in safety-critical applications, such as autonomous driving, where an adversarially altered ``Stop" sign could be misinterpreted as ``Go", potentially leading to catastrophic outcomes \citep{XU2023103143,LI2025104214,west2023towards}.

\textbf{Adversarial Defense and Detection}: To enhance the security and robustness of deep learning models against adversarial examples, \citep{r2} first introduced adversarial examples into the training process, pioneering the research into model robustness. This approach was later extended by studies showing that training models with adversarial examples generated via PGD attacks could significantly improve robustness \citep{mkadry2017towards}. Further advancements led to ensemble adversarial training, which incorporated transferable perturbations, further enhancing model robustness \citep{tramer2017ensemble}. The concept of Smooth Adversarial Training introduced a different perspective, achieving notable robustness improvements \citep{xie2020smooth}. Other methods, including self-supervised adversarial training \citep{naseer2020self} and causal parameter estimation \citep{lee2023mitigating}, further diversified the strategies to improve robustness. Additionally, researchers efforts extended to 3D point cloud recognition, highlighting the broad applicability of adversarial defense techniques \citep{ji2023benchmarking}. These studies have cumulatively enriched the adversarial training landscape.

As research progressed, the focus shifted toward optimizing the finer details of adversarial training. The analysis of smoothness enhancement revealed its specific role in domain adversarial training for improved target domain generalization \citep{rangwani2022closer}. Later, techniques to constrain loss variations between epochs and introduce convergence strides were developed to enhance adversarial training efficiency \citep{zhao2023fast}. Feature denoising methods emerged as a way to increase robustness by removing noise from input features during inference, although they faced limitations with resistance to white-box attacks and increased computational costs \cite{xie2019feature}. In response, selective feature regeneration was proposed as an alternative \citep{borkar2020defending}, while image restoration-based denoising was explored, though it was prone to content distortion and vulnerability to EOT attacks \citep{8844865}.

As adversarial attack strategies diversified, targeted defenses were developed. Image-semantic dual adversarial training (ISDAT) addressed the limited diversity of adversarial examples \citep{SUI2025110968}. For high-intensity perturbations, non-robust loss adjustment methods proved effective in improving model resilience \citep{10920873}. Advances in bidirectional mapping and self-attention feature alignment further bolstered resistance to attacks \citep{10890747}, while universal attention mechanisms successfully countered diverse attention-based adversarial attacks \citep{zhao2025universal}. These targeted strategies have significantly improved defenses across various attack scenarios.

In parallel, adversarial sample detection has become an increasingly critical area of research. Early methods focused on gradient masking to reduce the model’s sensitivity to small input perturbations \citep{papernot2017practical}. Subsequently, denoising-based reconstruction methods were developed to detect adversarial examples by identifying reconstruction errors and prediction discrepancies \citep{Meng}. Projection-based methods measured the prediction probability distance for sample detection \citep{xu2017feature}, while local intrinsic dimensionality analysis emerged as an effective technique \citep{ma2018characterizing}. Other approaches, including Mahalanobis distance \citep{NEURIPS2018_abdeb6f5}, natural scene statistics \citep{9206959}, and autoencoders \citep{sotgiu2020deep}, collectively established a robust adversarial detection framework.

Novel perspectives have further propelled advancements in the field. Key feature modulation frameworks solved adversarial patch detection \citep{wu2024napguard}, while clustering effect analysis explained adversarial robustness and introduced regularization techniques \citep{jin2023explaining}. Lightweight ensemble attacks (LEA2) presented new detection strategies for adversarial examples \citep{qian2023lea2}, and dual-function defense frameworks provided comprehensive mitigation for adversarial instances \citep{10942403}. Models such as Adversarial Surgery and Regeneration (ASR) significantly improved generalization and robustness \citep{10737447}.

\textbf{Masking Techniques and Entropy-Based Adversarial Attacks}: Masking techniques have emerged as essential tools in adversarial defense. Instance-binding augmentation techniques, for example, reconstructed perturbation distributions via masking branches \citep{10620338}. Gradient-based masking was shown to enhance attack transferability by perturbing sensitive regions \citep{zhang2024enhancing}. Information masking and region intersection strategies were developed to purify adversarial examples \citep{liu2025adversarial}, while in audio attacks, random masking optimized adversarial training processes \citep{bui2024boosting}. Variance-based masking (RAPID) was another approach for detecting candidate regions \citep{kim2024rapid}. These techniques underscore the growing importance of masking strategies in adversarial research.

From an information-theoretic perspective, high entropy in adversarial patches facilitates their localization and removal \citep{tarchoun2024information}. Entropy-based detectors have successfully identified adversarial examples by analyzing entropy differences before and after bit-depth reduction \citep{ryu2024detection}. Advanced entropy analysis further refined patch localization methods \citep{tarchoun2023jedi}, providing novel detection techniques for the field.

\section{Method}\label{sec3}

The fundamental principle of adversarial example detection lies in identifying and leveraging discriminative features to effectively distinguish adversarial examples from clean ones. This section first establishes the theoretical foundation and the motivation for this study, followed by a visualization-based analysis of the distinct vulnerability characteristics exhibited by adversarial examples in contrast to clean samples. Building on these observations, we introduce a quantitative metric, termed Sliding Mask Confidence Entropy (SMCE), to systematically assess the degree of adversarial vulnerability. To further elucidate the effectiveness of this metric, we visualize the quantified vulnerability distributions, thereby establishing a robust framework for distinguishing adversarial examples from their clean counterparts.

\subsection{Research Objective}\label{subsec1}

The central objective of this study is to enhance the security of artificial intelligence systems by enabling them to effectively resist adversarial attacks. Specifically, this work aims to uncover the inherent characteristics of adversarial examples from a novel perspective, and to leverage these characteristics to accurately classification between adversarial examples and clean samples, thereby mitigating the impact of adversarial attacks.

\subsection{Motivation and Intuition}\label{subsec1}

\begin{figure}
	\centering
	\includegraphics[width=1\linewidth]{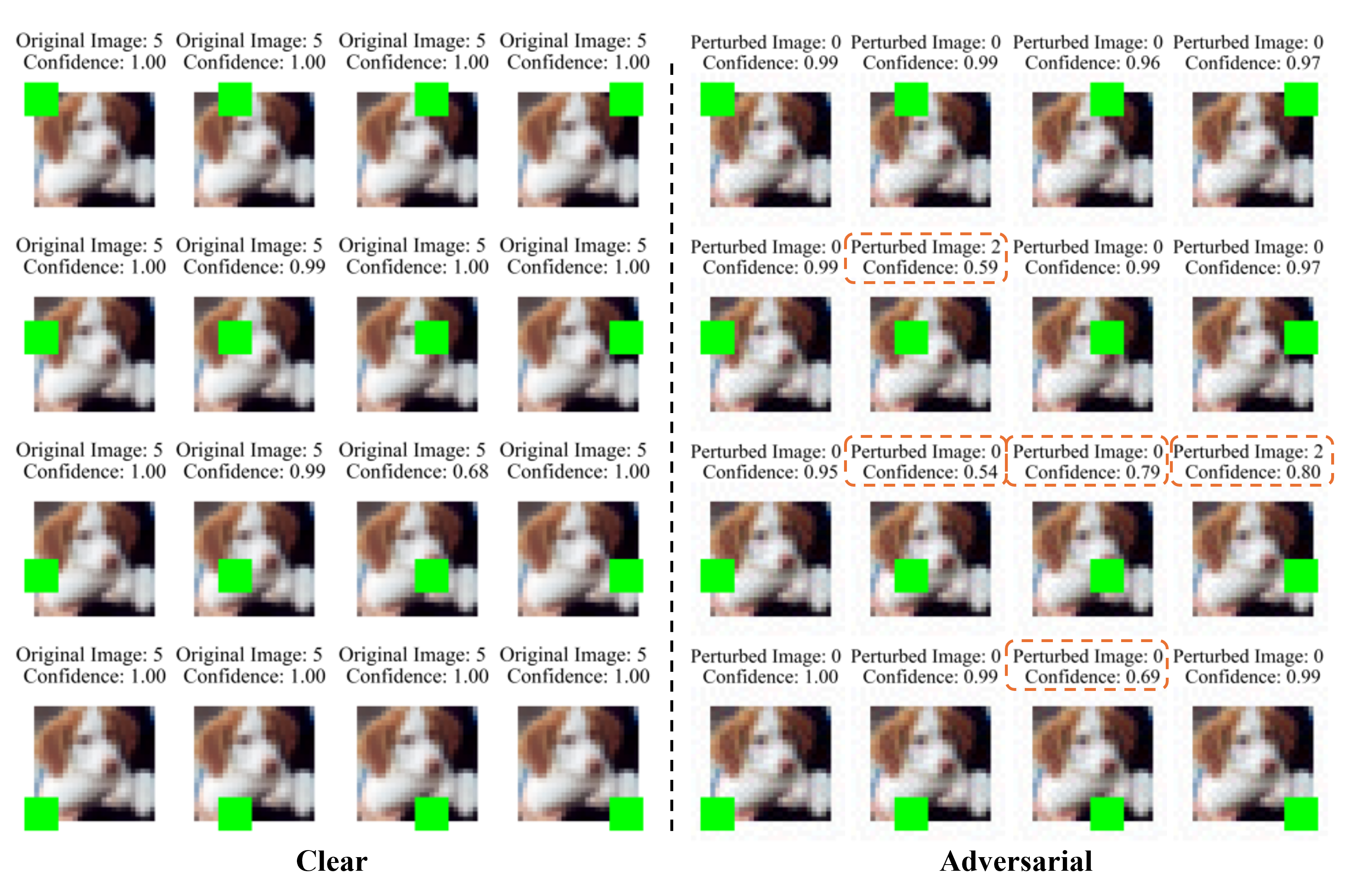}
	\caption{Seeking the vulnerability of adversarial examples}
	\label{fig:enter-label20}
\end{figure}

The goal of adversarial attacks is to induce misclassification in DNN models with minimal perturbations, which can be formally expressed as the following optimization problem:

\begin{equation}
	f(x) \neq f(x+r) \text{ subject to } r \leq \epsilon 
	\label{eq:refname2}
\end{equation}

Where \(f(x)\) denotes the predicted class for the original image, \(f(x+r)\) represents the predicted class for the adversarial example, and \(\epsilon\) is a small, predefined perturbation bound. Based on the preceding formulation, it follows that an adversarial example \(x+r\) arises from the superposition of a small perturbation \(r\) onto the original image \(x\). Although the magnitude of this perturbation is minimal, it is sufficient to induce misclassification in deep learning models. Mathematically, the adversarial example \(x+r\) can be viewed as a linear combination of the original image and the perturbation, underscoring the fundamental mechanism of adversarial attacks—where imperceptible perturbations in the input space lead to substantial alterations in the model’s output.

Further analysis suggests that the perturbation \(r\) can be interpreted as a specialized form of noise. Despite its constrained numerical range, its influence on the model’s classification outcomes is nontrivial. Related study reveals that the introduction of noise to an image not only perturbs its pixel value distribution but also compromises its stability \citep{tian2020deep,buades2005review}. We posit that adversarial examples exhibit increased vulnerability relative to their unperturbed counterparts, characterized by more ambiguous class boundaries that are inherently more prone to misclassification. This heightened vulnerability arises from the proximity of adversarial examples to decision boundaries in feature space, rendering them particularly sensitive to small input variations. 

Adversarial examples exhibit greater susceptibility to perturbations compared to clean samples. Building upon this foundation, we further investigate the underlying factors contributing to this vulnerability. Notably, the human visual system can reliably recognize objects even when images are partially occluded. Motivated by this observation, we examine how classification outcomes for clean and adversarial examples evolve when their occluded counterparts are fed into a deep neural network. 

\subsection{Vulnerability Exploration through Mask Box}\label{subsec1}
Drawing inspiration from human perception, we systematically apply an $m \times m$ dark occlusion block that traverses local regions of both clean and adversarial examples. By analyzing the resulting shifts in classification decisions, we aim to uncover the fundamental basis of adversarial vulnerability.

Figure~\ref{fig:enter-label20} presents a detailed analysis of the vulnerability of adversarial examples in the context of image classification. In this experiment, we introduced a occlusion block, referred to as the ``mask box", and used it to occlude various parts of both clean images and adversarial examples. The occlusion was applied systematically from left to right and top to bottom. We then utilized a ResNet-18 classifier to make predictions on the occluded images, with the aim of investigating the stability of clean images and adversarial examples under occlusion and their corresponding classification results. The 16 sub-figures on the left side of Figure~\ref{fig:enter-label20} illustrate the sliding of a $7 \times 7$ green mask box across the original sample. Each sub-figure presents the classification results, including the predicted class label and the confidence score for that label, as determined by the ResNet-18 classifier. Notably, when occluding the clean sample with the sliding mask, the model’s confidence and predicted class label remained largely unaffected. Even with partial occlusions, the clean image was consistently classified correctly, demonstrating the robustness of the model to such perturbations.

In contrast, the 16 sub-figures on the right side of Figure~\ref{fig:enter-label20} reveal a striking difference in behavior when adversarial examples are occluded. In these cases, the classification results exhibited significant instability, characterized by considerable drops in confidence scores. This observation underscores the inherent vulnerable of adversarial examples, which are highly susceptible to perturbations. This instability can be exploited as a key characteristic for detecting adversarial examples, offering a potential avenue for enhancing defenses against adversarial attacks.

Adversarial examples have been shown to exhibit significant instability, necessitating the development of a robust metric to quantify this instability. This paper introduces a novel formula to measure the variations in confidence scores and class labels resulting from occlusions in an image. By analyzing the differences in instability between clean samples and adversarial examples, we provide a clear methodology for distinguishing between these two categories. The key innovation lies in the introduction of a precise metric that enables effective differentiation between adversarial and clean samples, providing new insights into the behavior of adversarial attacks and advancing adversarial examples detection techniques.

\subsection{Sliding Mask Confidence Entropy}\label{subsec1}
This study addresses a fundamental challenge in adversarial example detection: quantifying how an image's classification confidence varies under partial occlusion. To systematically evaluate confidence stability, we propose Sliding Mask Confidence Entropy (SMCE), a novel metric that measures the uncertainty in classification confidence as an image undergoes successive occlusions via a sliding window mask. One of the key innovation of this work is the entropy-based formulation of SMCE, which enables precise quantification of confidence stability during occlusion.

The formula for Sliding Mask Confidence Entropy (SMCE) is as follows:

\begin{equation}
	H_{\text{SMCE}}(I) = \frac{1}{n} \sum_{i=1}^{n} \left( -\sum_{j=1}^{m} p_{ij} \log_2(p_{ij}) \right)
	\label{eq:SMCE}
\end{equation}
where \( H_{\text{SMCE}}(I) \) represents the Sliding Mask Confidence Entropy of image \( I \). \( p_{ij} \) is the confidence that image \( I \) is classified as the \( j \)-th class after being occluded by the \( i \)-th sliding window mask position.
\( n \) is the number of sliding positions of the window mask, i.e., the number of different positions the mask moves across the image.
\( m \) is the total number of classes that the classifier can identify, i.e., the number of possible categories for the image. 

The formula \ref{eq:SMCE} calculates the average of confidence entropy across all sliding window mask positions to assess the stability of the image's confidence during the occlusion process. Specifically, for each sliding position \( i \), we first compute the confidence entropy at that position, i.e., \(-\sum_{j=1}^{m} p_{ij} \log_2(p_{ij})\). Then, we sum up the confidence entropies for all positions and take the average to obtain the final Sliding Mask Confidence Entropy \( H_{\text{SMCE}}(I) \). A higher value of this metric indicates greater uncertainty in confidence changes under different occlusion conditions, implying poorer stability of the confidence; conversely, a lower value indicates more stable confidence changes, suggesting that the image's classification results are more reliable under various occlusion conditions.
\subsubsection{Property about SMCE}\label{subsubsec1}

\begin{property}
	\textbf{Non-negativity} .
	\begin{align}
		H_{\text{SMCE}}(I) \geq 0 \label{thm:theorem1}
	\end{align}
\end{property}
The Sliding Mask Confidence Entropy (SMCE) metric is inherently non-negative, as established in formula \ref{thm:theorem1}. When the SMCE value approaches zero, it indicates that the predicted labels remain consistent across diverse occlusion conditions, with model confidence converging towards one. This behaviour reflects high resilience to localized perturbations and suggests strong intrinsic image stability.

\begin{property}
	\textbf{Maximum Value of the SMCE} .
	\begin{align}
		H_{\text{SMCE}}(I) \leq \log_2 m \label{thm:theorem2}
	\end{align}
\end{property}

Furthermore, formula \ref{thm:theorem2} demonstrates that the maximum value of SMCE is \(\log_2 m\). As the image undergoes successive occlusions through sliding masks, greater variations in confidence scores and class labels lead to higher SMCE values. These elevated values signify reduced stability, emphasizing the image's vulnerability to adversarial examples. One of the key innovation of this work lies in quantitatively linking SMCE to the degree of stability.


\subsubsection{The Calculation Process of SMCE}\label{subsubsec1}
In the previous section, we introduced the Sliding Mask Confidence Entropy (SMCE) as a quantitative measure of image stability about adversarial or clean samples. This section provides a detailed exposition of the SMCE computation process.

\begin{figure}
	\centering
	\includegraphics[width=1.0\linewidth]{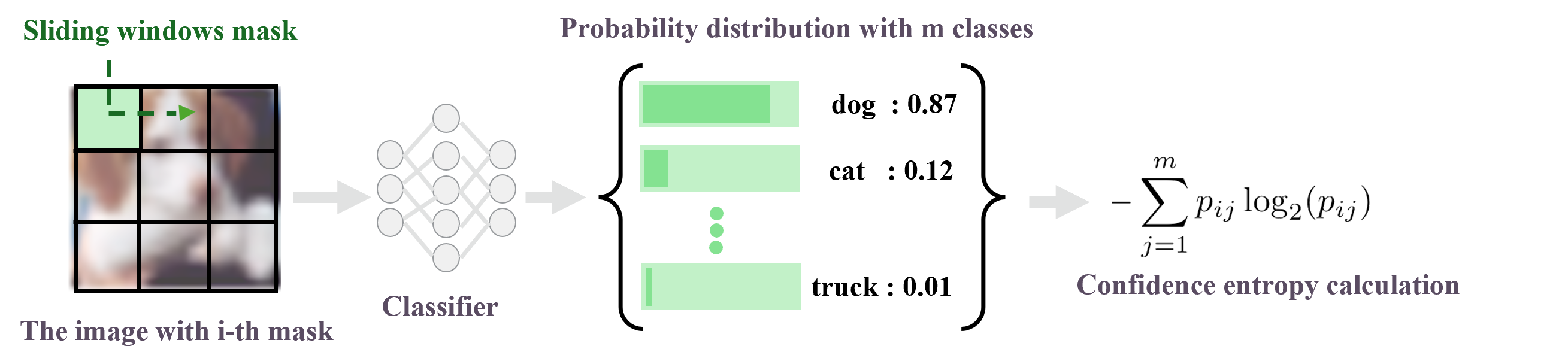}
	\captionsetup{justification=justified}
	\caption{The process of calculating confidence entropy for a single occluded image.}
	\label{fig:view2}
\end{figure}

When an image is classified through a deep learning model, the model outputs a classification vector and a classification label. As shown in the probability distribution diagram in Figure~\ref{fig:view2}, the classification vector represents the probability distribution output by the classifier, reflecting the confidence levels of the input image belonging to each predefined category. The classification vector is a numerical vector, with each element corresponding to a category, having a value range of [0, 1], and the sum of all elements equals 1. The classification label is the final classification result generated by the classifier based on the classification vector, indicating the most likely category to which the image belongs. Figure~\ref{fig:view2} illustrates the confidence entropy computation process for a single occluded image. Specifically, the image is first partially occluded using a mask block. The occluded image is then fed into the deep learning model for classification to obtain the classification vector and label. Subsequently, the classification vector is substituted into the confidence entropy calculation formula to obtain the confidence entropy of the image under the occluded state.

By sliding the mask block from left to right and top to bottom in sequence, and repeatedly computing the confidence entropy, the SMCE value of the image can be obtained by averaging all the computed results.

\subsubsection{Mask Entropy Field Map}\label{subsubsec1}

\begin{figure}
	\centering
	\includegraphics[width=0.5\linewidth]{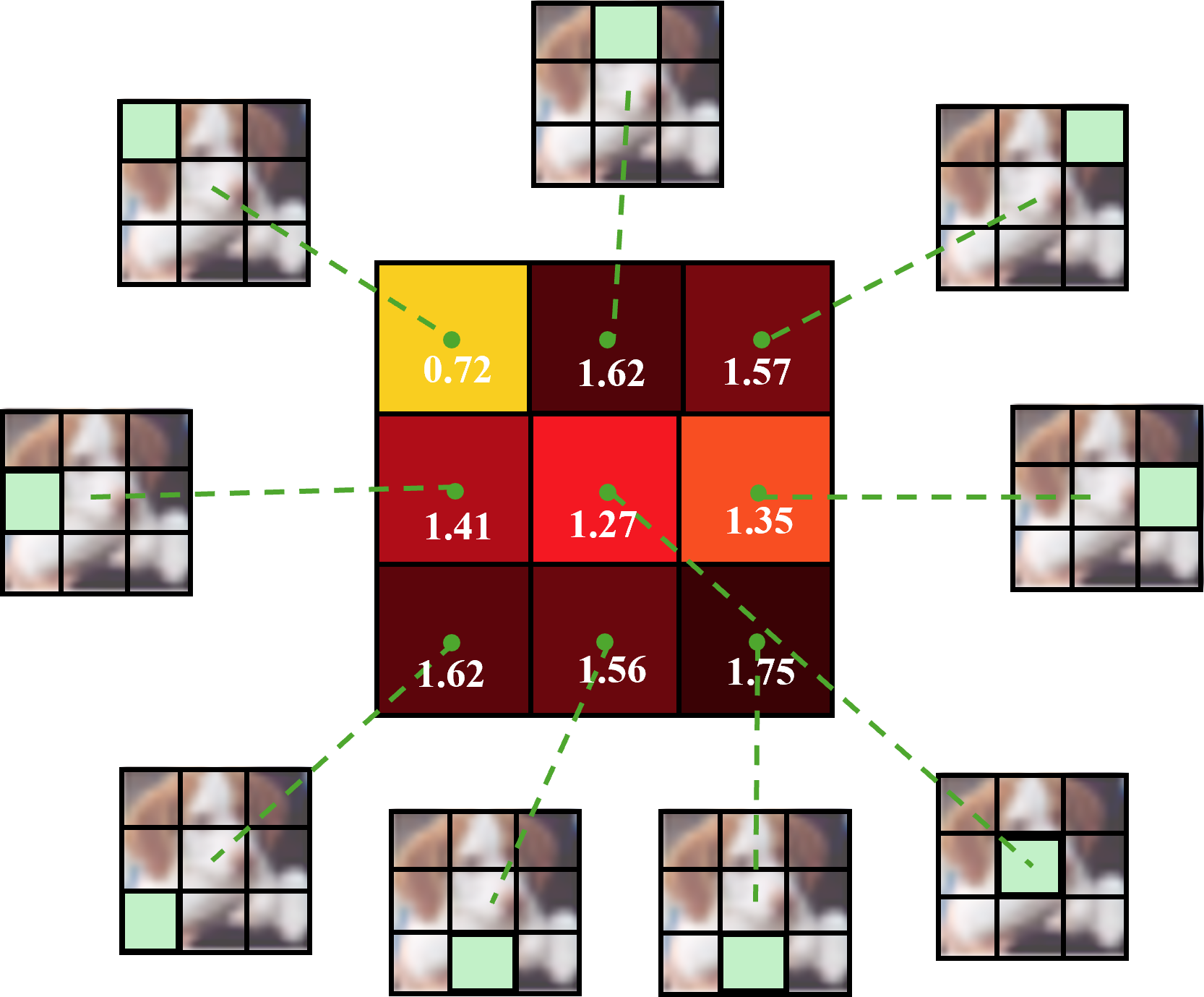}
	\captionsetup{justification=justified}
	\caption{Mask Entropy Field Map.}
	\label{fig:view3}
\end{figure}
The Sliding Mask Confidence Entropy (SMCE) is an indicator used to evaluate the stability of images under occlusion conditions. To more intuitively display the differences in stability between adversarial and clean samples, a visualization method called the Mask Entropy Field Map (MEFM) is proposed. As shown in Figure~\ref{fig:view3}, this method involves sliding a mask over the image to progressively occlude different regions and calculating the output confidence entropy of the classifier for each occluded image. Specifically, the entropy values at each position are mapped onto the image, where regions with higher entropy values (i.e., lower classifier confidence) are represented by warm colors (such as red and black), and regions with lower entropy values (i.e., higher classifier confidence) are represented by cool colors (such as yellow and white). The Mask Entropy Field Map is not only clearly demonstrates the changes in stability under different occlusion conditions but also intuitively reveals the significant differences in stability between adversarial and clean samples.

\begin{figure}
	\centering
	\includegraphics[width=1.0\linewidth]{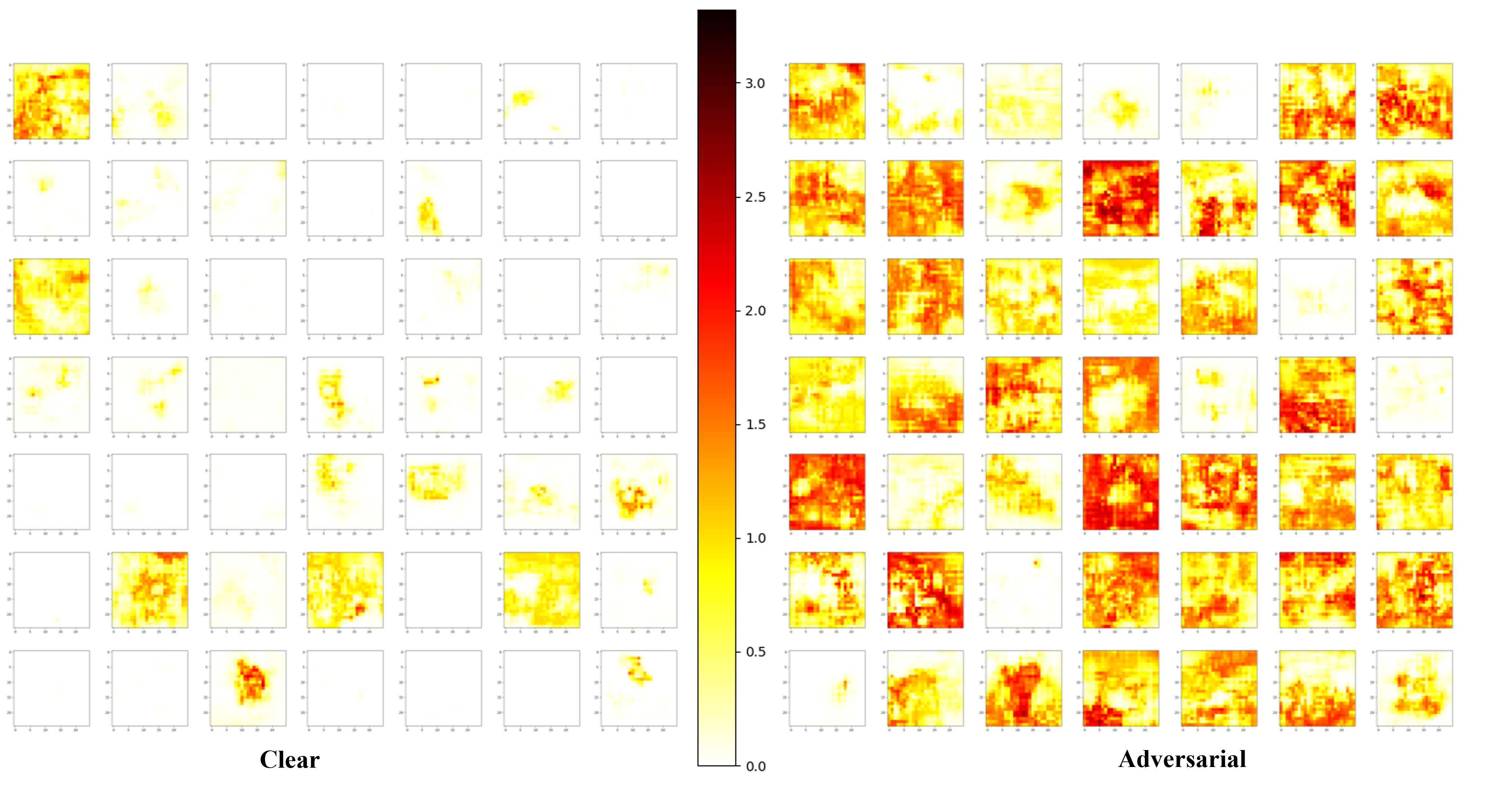}
	\captionsetup{justification=justified}
	\caption{Mask Entropy Field Map for clean samples and adversarial examples.}
	\label{fig:view4}
\end{figure}

To further elucidate the stability differences between adversarial examples and clean samples, a subset of clean samples from the CIFAR-10 dataset was randomly selected, and their Mask Entropy Field Map (MEFM) were generated, as shown in the ``Clear" subfigure of Figure~\ref{fig:view4}. Adversarial examples were then crafted from these clean samples using the Fast Gradient Sign Method (FGSM), and the corresponding MEFM were plotted, as illustrated in the ``Adversarial" subfigure of Figure~\ref{fig:view4}. The color intensity within the maps intuitively reflects image stability: darker regions correspond to higher Sliding Mask Confidence Entropy (SMCE) values, indicating reduced stability. A direct comparison between the two subfigures reveals that adversarial examples consistently exhibit darker MEFM, reflecting significantly lower stability compared to clean samples. This finding substantiates the inherent vulnerability of adversarial examples.

The presence of elevated entropy values in certain clean samples is likely due to underfitting in deep neural networks. While neural network classifiers achieve high performance on training datasets, attaining perfect classification accuracy remains challenging. Consequently, some clean samples are misclassified, thereby affecting the probability distribution in the SMCE computation and leading to higher entropy values. Conversely, the lower entropy values observed in some adversarial examples can be attributed to the loss of essential original features and the introduction of new category-related features induced by adversarial attack algorithms. This transformation enhances the robustness of the classifier against adversarial examples in the presence of occlusion, thereby yielding lower SMCE values. Experimental results reveal that adversarial examples consistently exhibit higher Sliding Mask Confidence Entropy (SMCE) values than clean samples. This distinct characteristic enables effective detection of adversarial examples through SMCE-based threshold.

\subsubsection{Adversarial Example Detection Algorithm}\label{subsubsec1}

\begin{figure}
	\centering
	\includegraphics[width=1\linewidth]{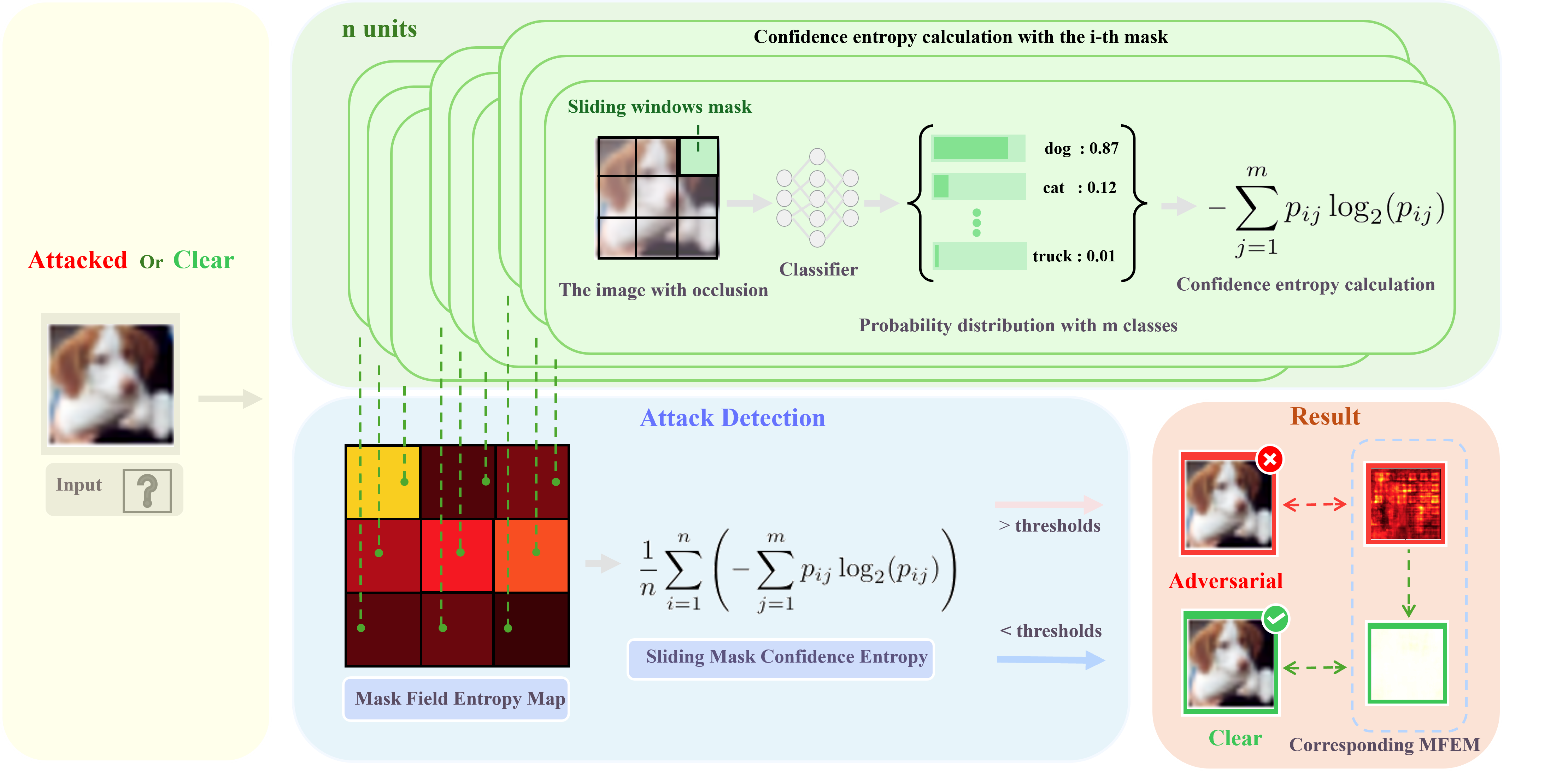}
	\captionsetup{justification=justified}
	\caption{The Sliding Window Masking-Adversarial Example Detection}
	\label{fig:view5}
\end{figure}

In the preceding analysis, the vulnerability of adversarial examples was systematically examined, revealing a marked decline in classification confidence and shifts in predicted class labels when subjected to occlusion via a sliding window. To quantitatively capture this behavior, we propose the Sliding Mask Confidence Entropy (SMCE) metric, which measures fluctuations in confidence scores and the frequency of class transitions under localized occlusion, thereby providing a precise quantitative assessment. Comparative analysis demonstrates that adversarial examples consistently yield significantly higher SMCE values than their clean counterparts, establishing a robust criterion for distinguishing adversarial examples from benign data.

Building on these findings, we propose an SMCE threshold-based approach for adversarial example detection. Given an input image, its SMCE value is first computed, if the value exceeds a predefined threshold, the image is identified as an adversarial example, prompting appropriate defensive measures, such as flagging or removing it from the dataset to mitigate its impact on model misclassification. Conversely, if the SMCE value falls below the threshold, the image is classified as a clean sample and can be retained for model training or inference. 

The proposed method for detecting adversarial examples, as depicted in Figure~\ref{fig:view5}, Upon inputting an image into the detection system, it is inherently uncertain whether the image is a pristine sample or one subjected to adversarial perturbations. To resolve this ambiguity, a novel strategy is employed: sliding masks are sequentially applied to occlude portions of the image, moving from the top-left corner to the bottom-right corner. Each occluded version of the image is then processed through a deep learning model, yielding a key metric—Sliding Mask Confidence Entropy (SMCE). The SMCE quantifies the stability of the image’s classification under occlusion, with a low value indicating high confidence that the image is a clean sample. Conversely, a high SMCE value signals instability in classification, suggesting the presence of adversarial manipulations. By setting a predefined threshold for SMCE, the method effectively classifies images as either unaltered or adversarial. Algorithm~\ref{fig:adversarial_detection} introduces a crucial innovation in adversarial detection, as it enables precise identification of adversarial examples by leveraging occlusion-induced uncertainty, providing a robust approach to safeguard against adversarial attacks in real-world applications.

\begin{algorithm}[H]
	\caption{SWM-AED: The Sliding Window Masking-Adversarial Example Detection}
	\begin{algorithmic}[1]
		\Procedure{SWM-AED}{$x, f, m, \text{threshold}$}
		\State \textbf{Input:} Image $x$, classifier $f$, mask size $m$, threshold $0.1$ (default).
		\State \textbf{Output:} Boolean indicating if $x$ is an adversarial example.
		\State \textbf{Initialize:} $H_{\text{SMCE}} \gets 0$, $n \gets \text{number of windows}$.
		\For{each sliding window mask $M_i$ in image $x$}
		\State $p_{ij} = f_j(x \odot M_i)$
		\State $H_{\text{}} = -\sum_{j=1}^{m} p_{ij} \log_2(p_{ij})$
		\State $H_{\text{SMCE}} \gets H_{\text{SMCE}} + H_{\text{}}$
		\EndFor
		\State $H_{\text{SMCE}} \gets \frac{H_{\text{SMCE}}}{n}$
		\If{$H_{\text{SMCE}} > \text{threshold}$}
		\State \Return True \Comment{Adversarial example}
		\Else
		\State \Return False \Comment{Not an adversarial example}
		\EndIf
		\EndProcedure
	\end{algorithmic}\label{fig:adversarial_detection}
\end{algorithm}
where $x \odot M_i$ denotes the region-wise occlusion of the image $x$ by the $i$-th sliding window mask $M_i$ of size $m \times m$, and $p_{ij}$ denotes the probability score assigned by the classifier $f$ when predicting that the image $x$ belongs to class $j$, and $x$ is occluded by the $i$-th sliding mask of size $m \times m$.

\section{Experimental}\label{sec4}
\subsection{Datasets and Classifiers}
All experiments are conducted on the widely adopted CIFAR-10~\citep{krizhevsky2009learning}, which comprises 60,000 colour images ($32 \times 32$ pixels) evenly distributed across ten object categories. Its balanced class distribution and moderate image resolution make it a standard benchmark in adversarial examples research. In the comparative experiments that reveal the sensitivity of adversarial examples to occlusion (e.g., Figure~\ref{fig:view4} and Figure~\ref{fig:enter-label1}), we randomly selected 1,800 images from the 10,000-image CIFAR-10 test set to form the control group. For the experimental group, we generated corresponding adversarial examples from these clean images using nine different adversarial attack algorithms, with each algorithm producing 200 adversarial samples. In the evaluation of the detection algorithm, we combined the experimental group (adversarial examples) and the control group (clean images), and applied the SWM-AED detection method to assess the accuracy of adversarial example detection.

To comprehensively assess adversarial susceptibility, multiple deep neural network architectures with varying levels of accuracy—including ResNet-18, ResNet-50 and VGG-11—are trained on CIFAR-10. These models serve both as baselines and as entropy calculators for the proposed Sliding Window Masking-Adversarial Example Detection (SWM-AED) framework, which introduces the Sliding Mask Confidence Entropy (SMCE) as a discriminative feature for adversarial detection. Through extensive comparative experiments, SWM-AED consistently outperforms conventional defense methods across multiple evaluation metrics. These results highlight its robustness, scalability and potential as a generalizable solution to adversarial vulnerability in deep learning models.


\subsection{Evaluation Metrics}\label{subsec1}

The SWM-AED algorithm we proposed can be regarded as a binary classifier, used to distinguish between adversarial examples and clean samples. In this context, the positive class represents adversarial examples, while the negative class represents clean samples that have not been attacked. We use Precision, Recall and F1 score as Three important metrics to evaluate the performance of this adversarial example detection algorithm. 

Confusion Matrix: A confusion matrix is a table used to evaluate the performance of a classification model. It contains four key components,
TP (True Positive): Correctly predicted as adversarial.
FP (False Positive): Incorrectly predicted as adversarial.
FN (False Negative): Incorrectly predicted as clean.
TN (True Negative): Correctly predicted as clean.

For binary classification problem, the confusion matrix can be visualized as follows:

\begin{table}[ht]
	\centering
	\caption{Confusion Matrix}
	\label{tab:confusion_matrix}
	\begin{tabular}{lcc}
		\specialrule{1.2pt}{0pt}{2pt} 
		\diagbox{Actual}{Predicted} & \textbf{Adversarial (Positive)} & \textbf{Clean (Negative)} \\
		\midrule
		\textbf{Adversarial}      & TP & FN \\
		\textbf{Clean}            & FP & TN \\
		\bottomrule
	\end{tabular}
\end{table}

Precision: Precision is the proportion of samples that are actually adversarial among those predicted as adversarial. The formula for calculating Precision is:

\[ \text{Precision} = \frac{\text{TP}}{\text{TP} + \text{FP}} \]

Recall: Recall is the proportion of samples that are correctly predicted as adversarial among all actual adversarial examples. It reflects the extent to which the algorithm can detect adversarial examples. The formula for calculating Recall is:

\[ \text{Recall} = \frac{\text{TP}}{\text{TP} + \text{FN}} \]

Accuracy: Accuracy denoted as Acc, is a widely used and intuitive evaluation metric. It represents the proportion of correctly predicted samples out of the total number of samples. This metric provides a comprehensive reflection of the overall prediction accuracy of a model.

\[
\text{Acc} = \frac{\text{TP + TN}}{\text{TP + TN + FP + FN}}
\]  

F1 Score: The F1 score is the harmonic mean of Precision and Recall, providing a single metric that balances both. The formula for calculating the F1 score is:

\[ \text{F1} = 2 \times \frac{\text{Precision} \times \text{Recall}}{\text{Precision} + \text{Recall}} \]

This score is particularly useful when the class distribution is unbalanced, as it gives a more balanced view of the model's performance.

\subsection{Experimental Results}
In this experiment, the vulnerability of adversarial examples is comprehensively investigated, with particular emphasis on analyzing the distribution of SMCE values for adversarial examples generated using various attack methods. These distributions are systematically compared to those of clean samples. To further enhance the generalizability of the SWM-AED algorithm, we explore the impact of threshold settings on its performance, ensuring its efficacy across a wide range of adversarial attacks. Additionally, the influence of key factors (including mask size, model accuracy, model depth, and architecture) on the detection performance of the SWM-AED algorithm is thoroughly examined. Finally, comparative experiments with existing state-of-the-art adversarial defense algorithms demonstrate the superior detection accuracy of the proposed approach, validating its significant advantages.

\subsubsection{The Empirical Distribution of the Sliding Mask Confidence Entropy}
\begin{figure}
	\centering
	\includegraphics[width=1\linewidth]{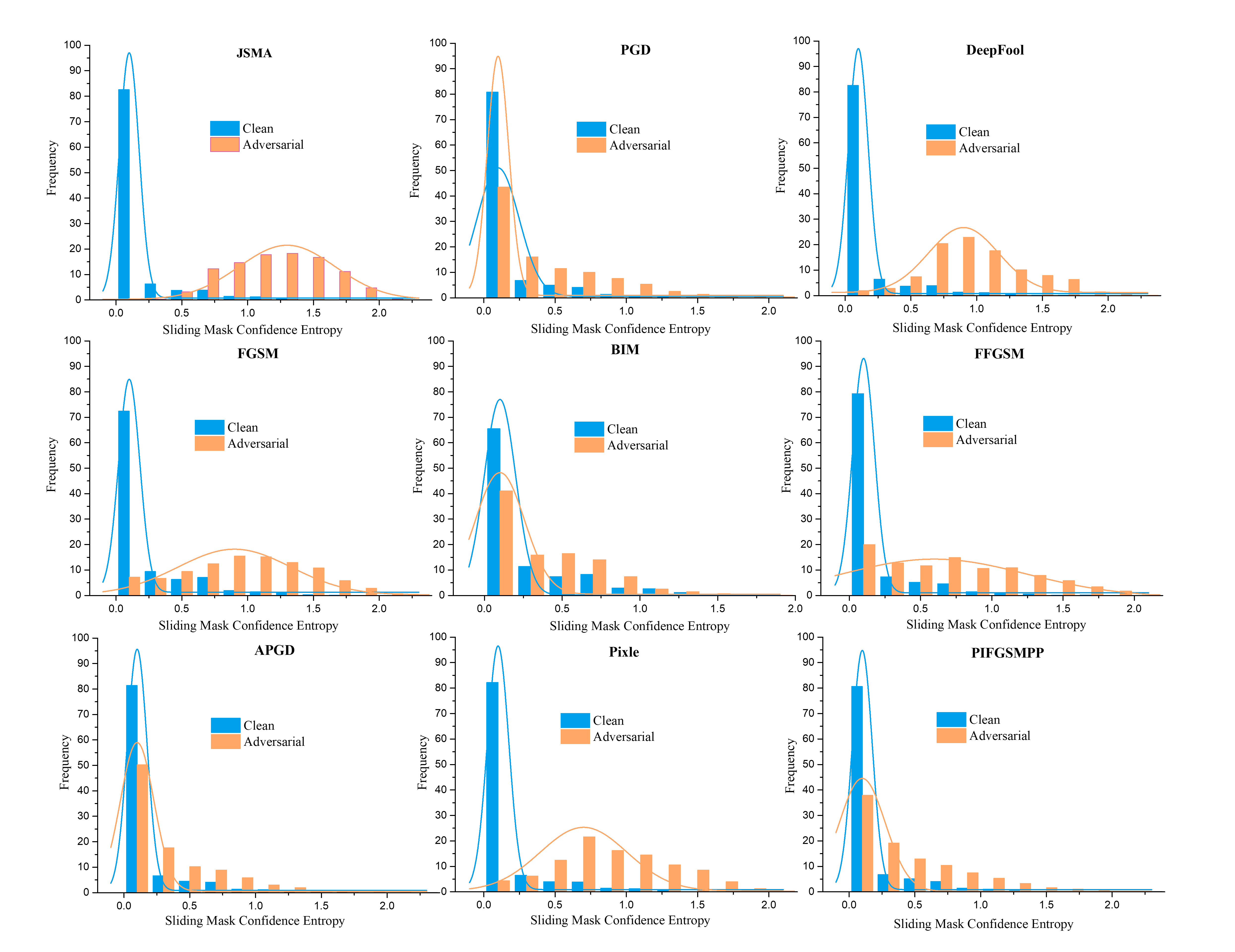}
	\caption{The figure presents a comparative analysis of the Sliding Mask Confidence Entropy distributions for adversarial examples generated by nine distinct attack algorithms and their original, unperturbed counterparts. Empirical distributions are visualized, while fitted Gaussian distributions are overlaid with line graphs. Specifically, yellow bars and lines represent adversarial examples, whereas blue bars and lines denote clean samples. The x-axis corresponds to the range of Sliding Mask Confidence Entropy values, which reflect the uncertainty in classification outcomes for image samples under occlusion. The y-axis indicates the statistical  frequency of Sliding Mask Confidence Entropy, providing insight into the density distribution across varying entropy levels.}
	\label{fig:enter-label1}
\end{figure}

\begin{figure}
	\centering
	\includegraphics[width=1.0\linewidth]{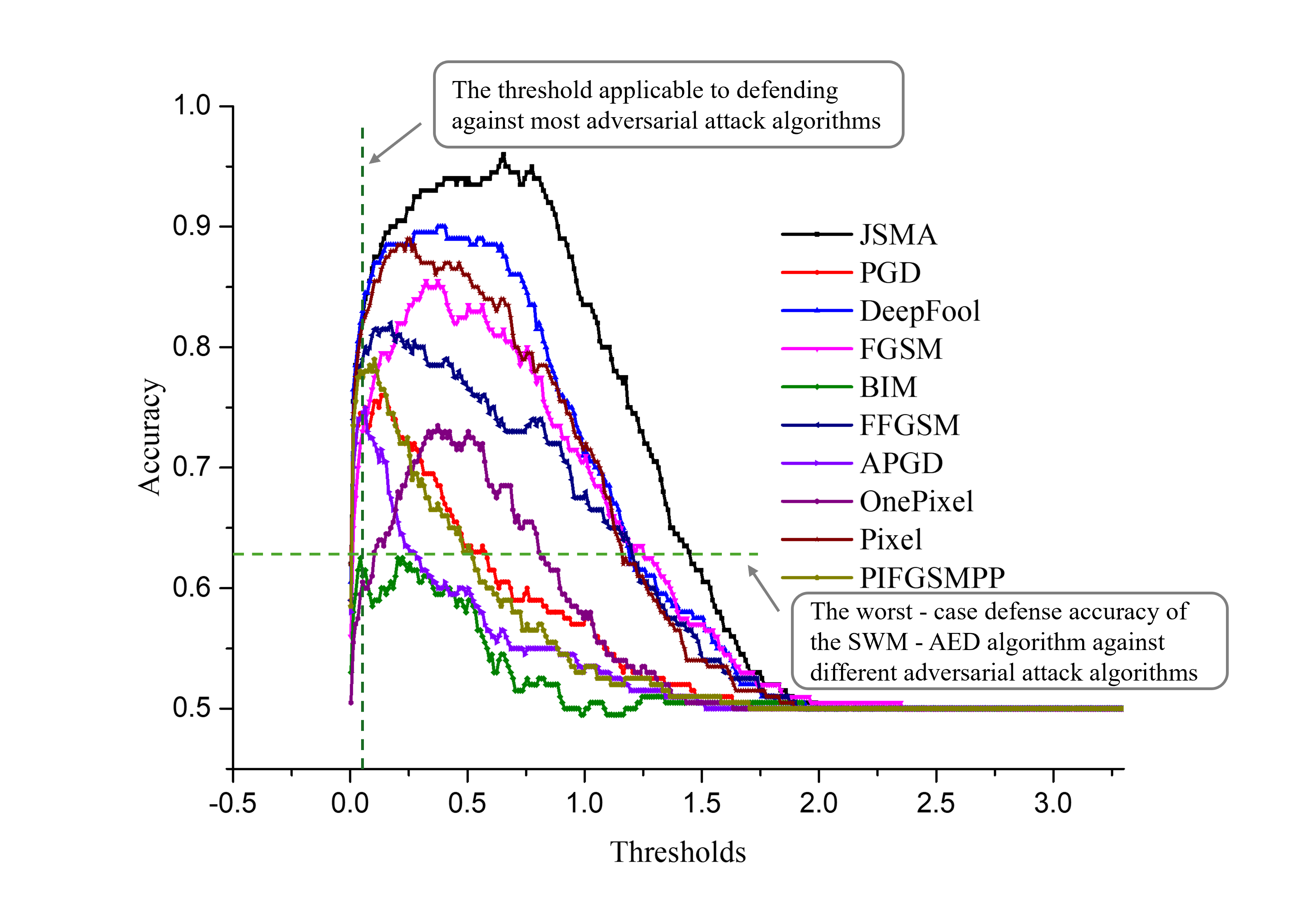}
	\caption{The figure illustrates the detection performance of the proposed method against adversarial examples generated by multiple attack algorithms. To ensure statistical significance, each algorithm generates a large-scale adversarial example dataset. The x-axis represents the range of classification decision thresholds, while the y-axis depicts the accuracy of the SWM-AED.}
	\label{fig:enter-label2}
\end{figure}

This study systematically evaluates nine widely used adversarial attack algorithms—JSMA\citep{7467366}, PGD\citep{mkadry2017towards}, DeepFool\citep{r4}, FGSM\citep{r2}, BIM\citep{kurakin2018adversarial}, FFGSM\citep{wong2020fast}, APGD\citep{croce2020reliable}, Pixel\citep{pomponi2022pixle}, and PIFGSMPP\citep{gao2020patch}—by analyzing the Gaussian empirical distributions of Sliding Mask Confidence Entropy for adversarial examples and their original, unperturbed counterparts. The results reveal notable differences in the confidence entropy distributions between adversarial and clean samples, which are visually distinguishable. As can be seen from Figure~\ref{fig:enter-label1}, yellow distributions correspond to adversarial examples, while blue distributions represent clean samples. Differences in both the position and shape of these distributions serve as indicators of adversarial robustness.  

The findings demonstrate a significant correlation between the magnitude of distributional differences and the robustness of adversarial examples. When the entropy distribution of adversarial examples deviates substantially from that of clean samples, the adversarial examples tend to be more fragile, exhibiting higher susceptibility to perturbations and greater instability in classification outcomes. Conversely, smaller distributional discrepancies indicate that the generated adversarial examples yield more stable classification results. As illustrated in Figure~\ref{fig:enter-label1}, among the nine tested algorithms, adversarial examples generated by BIM, PGD, APGD, and PIFGSMPP exhibit entropy distributions that closely resemble those of clean samples, suggesting enhanced stability. This similarity in entropy profiles indicates that adversarial examples produced by these four algorithms share characteristics more akin to normal, non-adversarial data. Consequently, these adversarial examples are more stable, as they are better able to evade detection by the SWM-AED algorithm when compared to samples generated by other attack methods in the evaluation.

Notably, the study finds that complete overlap between the distributions of adversarial and clean samples is exceedingly rare. This observation aligns with the fundamental nature of adversarial example generation: adversarial perturbations introduce deliberate noise into the original samples, inevitably disrupting their statistical properties and shifting their distributions.  

The entropy distribution plots provide direct insight into the vulnerability of adversarial examples. By examining these distributions, researchers can systematically assess the susceptibility of adversarial examples generated by different attack algorithms and further explore the impact of various adversarial attacks on model confidence.

\subsubsection{The Relationship between Threshold and Model Performance}

Since the proposed detection algorithm is inherently a binary classification task, the selection of an appropriate decision threshold is crucial for effectively distinguishing between adversarial and clean samples. The choice of this threshold directly influences detection performance. To systematically assess the optimal detection accuracy across different threshold settings, the experiment evaluates ten representative adversarial attack algorithms, including FGSM, PGD, and DeepFool, among others. This comprehensive analysis aims to rigorously examine the effectiveness of the detection method against a wide range of adversarial attacks.  

Figure~\ref{fig:enter-label2} depicts the variation in accuracy across different threshold conditions. The Sliding Mask Confidence Entropy (SMCE) values are computed using the ResNet-18 model to systematically assess the classification-based detection performance of the proposed SWM-AED algorithm across a range of threshold settings. Experimental findings reveal that SWM-AED consistently delivers strong detection performance against a wide spectrum of adversarial attacks. Notably, when the detection threshold is set to 0.1, the algorithm achieves robust generalisation, attaining detection accuracies exceeding 75\% for the majority of adversarial examples—excluding only the BIM and OnePixel attacks.
In particular, the detector demonstrates exceptional resilience against targeted attacks such as JSMA and DeepFool, achieving detection accuracies above 90\%. Although performance against BIM-like attacks—characterised by iterative gradient-based perturbation—is comparatively lower, the algorithm still maintains detection rates above 60\%, underscoring its robustness even in challenging scenarios. A comprehensive analysis of the detection results reveals that SWM-AED achieves an average detection accuracy exceeding 80\% across randomly sampled adversarial examples. Among the nine evaluated attack methods, two yield detection rates above 90\% and three exceed 80\%, highlighting the algorithm’s strong generalisation capability and reliable performance across diverse adversarial threat models.


\subsubsection{Investigation of Key Factors Influencing Algorithm Performance}
\label{sec:4.4.3}

\begin{table}[h!]
	\centering
	\caption{Based on the ResNet-18 model with a classification accuracy rate of 96\%, the mask detection method based on a $3 \times 3$ sliding window is used to evaluate its detection success rate for samples generated by various adversarial attack algorithms.}
	\label{tab:attack_metrics1}
	\begin{tabular}{@{}lcccc@{}}
		\toprule
		\textbf{Attack Method} & \textbf{Precision (\%)} & \textbf{Recall (\%)} & \textbf{F1 Score } & \textbf{Accuracy (\%)} \\ \midrule
		JSMA & 96.04 & 97.00 & 0.96 & 96.50 \\
		PGD & 71.43 & 70.00 & 0.70 & 71.00 \\
		DeepFool & 93.68 & 0.89 & 0.91 & 91.50 \\
		FGSM & 76.56 & 98.00 & 0.85 & 84.00 \\
		BIM & 61.54 & 56.00 & 0.58 & 60.50 \\
		FFGSM & 79.05 & 83.00 & 0.80 & 80.50 \\
		APGD & 64.66 & 75.00 & 0.69 & 67.00 \\
		OnePixel & 77.27 & 68.00 & 0.72 & 74.00 \\
		Pixle & 87.04 & 94.00 & 0.90 & 90.00 \\
		PIFGSMPP & 67.97 & 87.00 & 0.76 & 73.00 \\ \bottomrule
	\end{tabular}
\end{table}

\begin{table}[h!]
	\centering
	\caption{Based on the ResNet-18 model with a classification accuracy rate of 96\%, the mask detection method based on a $7 \times 7$ sliding window is used to evaluate its detection success rate for samples generated by various adversarial attack algorithms.}
	\label{tab:attack_metrics2}
	\begin{tabular}{@{}lccccccc@{}}
		\toprule
		\textbf{Attack Method} & \textbf{Precision (\%)} & \textbf{Recall (\%)} & \textbf{F1 Score}     & \textbf{Accuracy (\%)} \\ \midrule
		JSMA          & 95.10          & 97.00       & 0.96         & 96.00         \\
		PGD           & 76.53          & 75.00       & 0.75         & 76.00         \\
		DeepFool      & 86.36          & 95.00       & 0.90         & 90.00         \\
		FGSM          & 80.87          & 93.00       & 0.86         & 85.50         \\
		BIM           & 59.12          & 81.00       & 0.68         & 62.50         \\
		FFGSM         & 79.63          & 86.00       & 0.82         & 82.00         \\
		APGD          & 71.93          & 82.00       & 0.76         & 75.00         \\
		OnePixel      & 71.96          & 77.00       & 0.74         & 73.50         \\
		Pixle         & 84.21          & 96.00       & 0.89         & 89.00         \\
		PIFGSMPP      & 76.85          & 83.00       & 0.79         & 79.00         \\ \bottomrule
	\end{tabular}
\end{table}

\begin{table}[h!]
	\centering
	\caption{Based on the ResNet-18 model with a classification accuracy rate of 96\%, the mask detection method based on a $9 \times 9$ sliding window is used to evaluate its detection success rate for samples generated by various adversarial attack algorithms.}
	\label{tab:attack_metrics3}
	\begin{tabular}{@{}lcccc@{}}
		\toprule
		\textbf{Attack Method} & \textbf{Precision (\%)} & \textbf{Recall (\%)} & \textbf{F1 Score } & \textbf{Accuracy (\%)} \\ \midrule
		JSMA          & 88.39          & 99.00       & 0.93         & 93.00         \\
		PGD           & 71.30          & 82.00       & 0.76         & 74.50         \\
		DeepFool      & 82.76          & 96.00       & 0.88         & 88.00         \\
		FGSM          & 79.13          & 91.00       & 0.84         & 83.50         \\
		BIM           & 60.19          & 62.00       & 0.61         & 60.50         \\
		FFGSM         & 72.87          & 94.00       & 0.82         & 79.50         \\
		APGD          & 67.65          & 92.00       & 0.77         & 74.00         \\
		OnePixel      & 68.18          & 75.00       & 0.71         & 70.00         \\
		Pixle         & 83.18          & 89.00       & 0.85         & 85.50         \\
		PIFGSMPP      & 69.85          & 95.00       & 0.80         & 77.00         \\ \bottomrule
	\end{tabular}
\end{table}

\begin{table}[h!]
	\centering
	\caption{Based on the ResNet-18 model with a classification accuracy rate of 80.8\%, the mask detection method based on a $7 \times 7$ sliding window is used to evaluate its detection success rate for samples generated by various adversarial attack algorithms.}
	\label{tab:attack_metrics4}
	\begin{tabular}{@{}lcccc@{}}
		\toprule
		\textbf{Attack Method} & \textbf{Precision (\%)} & \textbf{Recall (\%)} & \textbf{F1 Score } & \textbf{Accuracy (\%)} \\ \midrule
		JSMA          & 73.33 & 88.00 & 0.80 & 78.00 \\
		PGD           & 58.70 & 81.00 & 0.68 & 62.00 \\
		DeepFool      & 67.94 & 89.00 & 0.77 & 73.50 \\
		FGSM          & 60.87 & 84.00 & 0.70 & 65.00 \\
		BIM           & 59.09 & 13.00 & 0.21 & 52.00 \\
		FFGSM         & 59.40 & 79.00 & 0.67 & 62.50 \\
		APGD          & 57.96 & 91.00 & 0.70 & 62.50 \\
		OnePixel      & 53.05 & 87.00 & 0.65 & 55.00 \\
		Pixle         & 68.09 & 96.00 & 0.79 & 75.50 \\
		PIFGSMPP      & 60.31 & 79.00 & 0.68 & 63.50 \\ \bottomrule
	\end{tabular}
\end{table}

\begin{table}[h!]
	\centering
	\caption{Based on the ResNet-50 model with a classification accuracy rate of 79.1\%, the mask detection method based on a $7 \times 7$ sliding window is used to evaluate its detection success rate for samples generated by various adversarial attack algorithms.}
	\label{tab:attack_metrics5}
	\begin{tabular}{@{}lcccc@{}}
		\toprule
		\textbf{Attack Method} & \textbf{Precision (\%)} & \textbf{Recall (\%)} & \textbf{F1 Score} & \textbf{Accuracy (\%)} \\ \midrule
		JSMA          & 78.07 & 89.00 & 0.83 & 82.00 \\
		PGD           & 67.16 & 90.00 & 0.76 & 73.00 \\
		DeepFool      & 70.87 & 90.00 & 0.79 & 76.50 \\
		FGSM          & 72.36 & 89.00 & 0.79 & 77.50 \\
		BIM           & 50.81 & 94.00 & 0.65 & 51.50 \\
		FFGSM         & 69.23 & 90.00 & 0.78 & 75.00 \\
		APGD          & 67.54 & 77.00 & 0.71 & 70.00 \\
		OnePixel      & 55.00 & 11.00 & 0.18 & 51.00 \\
		Pixle         & 61.74 & 92.00 & 0.73 & 67.50 \\
		PIFGSMPP      & 70.25 & 85.00 & 0.76 & 74.50 \\ \bottomrule
	\end{tabular}
\end{table}

\begin{table}[h!]
	\centering
	\caption{Based on the Vgg-11 model with a classification accuracy rate of 81.3\%, the mask detection method based on a $7 \times 7$ sliding window is used to evaluate its detection success rate for samples generated by various adversarial attack algorithms.}
	\label{tab:attack_metrics6}
	\begin{tabular}{@{}lcccc@{}}
		\toprule
		\textbf{Attack Method} & \textbf{Precision (\%)} & \textbf{Recall (\%)} & \textbf{F1 Score} & \textbf{Accuracy (\%)} \\ \midrule
		JSMA          & 73.68 & 98.00 & 0.84 & 81.50 \\
		PGD           & 73.28 & 85.00 & 0.78 & 77.00 \\
		DeepFool      & 75.63 & 90.00 & 0.82 & 80.50 \\
		FGSM          & 73.28 & 85.00 & 0.78 & 77.00 \\
		BIM           & 54.78 & 63.00 & 0.58 & 55.50 \\
		FFGSM         & 69.17 & 92.00 & 0.78 & 75.50 \\
		APGD          & 71.68 & 81.00 & 0.76 & 74.50 \\
		OnePixel      & 54.23 & 77.00 & 0.63 & 56.00 \\
		Pixle         & 73.58 & 78.00 & 0.75 & 75.00 \\
		PIFGSMPP      & 68.46 & 89.00 & 0.77 & 74.00 \\ \bottomrule
	\end{tabular}
\end{table}

From the analysis of formula~\ref{eq:SMCE}, it is evident that the SWM-AED algorithm requires a well-performing deep learning model to classify occluded images and generate the corresponding class probability distributions. These distributions are then used to calculate the SMCE, which is subsequently compared to a predefined threshold to detect and classify adversarial examples. The performance of the algorithm is primarily influenced by two key factors: the mask size and the intrinsic classification capability of the model. To systematically assess the impact of these factors on detection performance, we conduct a series of comparative experiments. These experiments examine the effects of mask size, classification accuracy of deep learning models, model architecture (including depth), and the generalization ability of the SWM-AED algorithm under various adversarial attacks.


\begin{figure}
	
	\centering
	\includegraphics[width=1\linewidth]{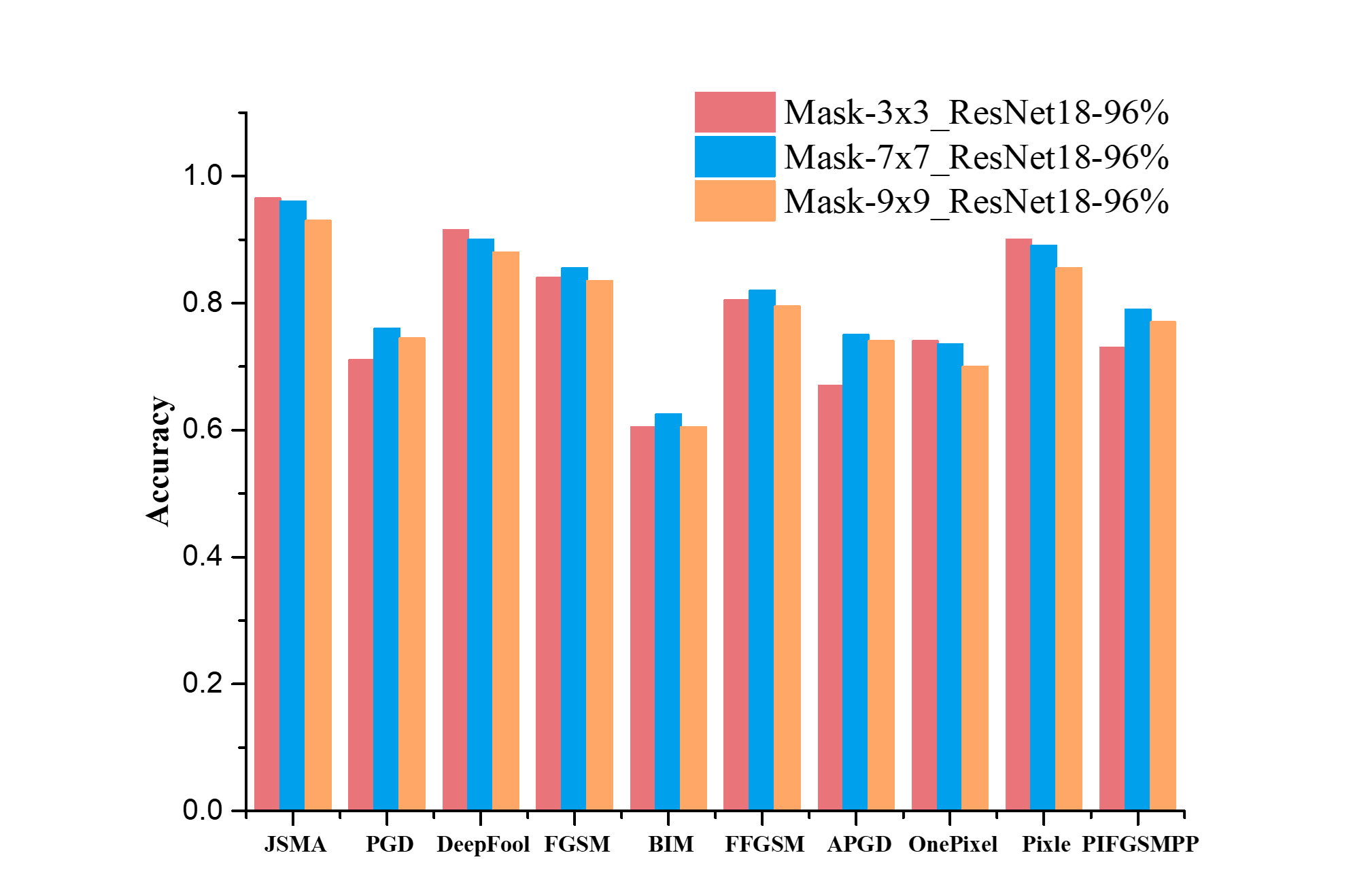}
	\captionsetup{justification=justified}
	\caption{Detection accuracy of the SWM-AED algorithm under various adversarial attack methods using different mask window sizes.}
	
	\label{fig:view10}
\end{figure}

\noindent\textbf{The impact of mask window size on the SWM-ADE detection algorithm.}
Experiments were performed using the same model with three different mask sizes—$3 \times 3$, $7 \times 7$, and $9 \times 9$—while evaluating their effects on detection accuracy. The corresponding results are presented in Tables~\ref{tab:attack_metrics1}, \ref{tab:attack_metrics2}, and \ref{tab:attack_metrics3}. The evaluation involved ten distinct adversarial attack algorithms and use four key performance metrics, including Precision, Recall, F1-score and Accuracy, for a comprehensive evaluation. For each attack algorithm, multiple classification thresholds were tested, and the optimal results were recorded.  

By synthesizing and analyzing the data presented in Tables~\ref{tab:attack_metrics1}, \ref{tab:attack_metrics2}, and \ref{tab:attack_metrics3}, Figure~\ref{fig:view10} is constructed. The figure distinctly illustrates substantial variations in the detection accuracy of the SWM-AED algorithm across different mask sizes. The experimental findings underscore the necessity of dynamically adjusting the mask size to optimize detection performance for various adversarial attack strategies. Further comparative analysis reveals that the SWM-AED algorithm attains optimal performance in most adversarial example detection tasks when the mask size is configured to $7 \times 7$. This finding suggests that neither the smallest nor the largest mask necessarily yields optimal results; rather, the choice of mask size should be tailored to the image dimensions and the specific characteristics of the detection task. A mask that is too small may fail to effectively occlude local image features, reducing its ability to capture adversarial perturbations, whereas an excessively large mask may introduce unnecessary noise, compromising local feature detection.


\begin{figure}
	
	\centering
	\includegraphics[width=1\linewidth]{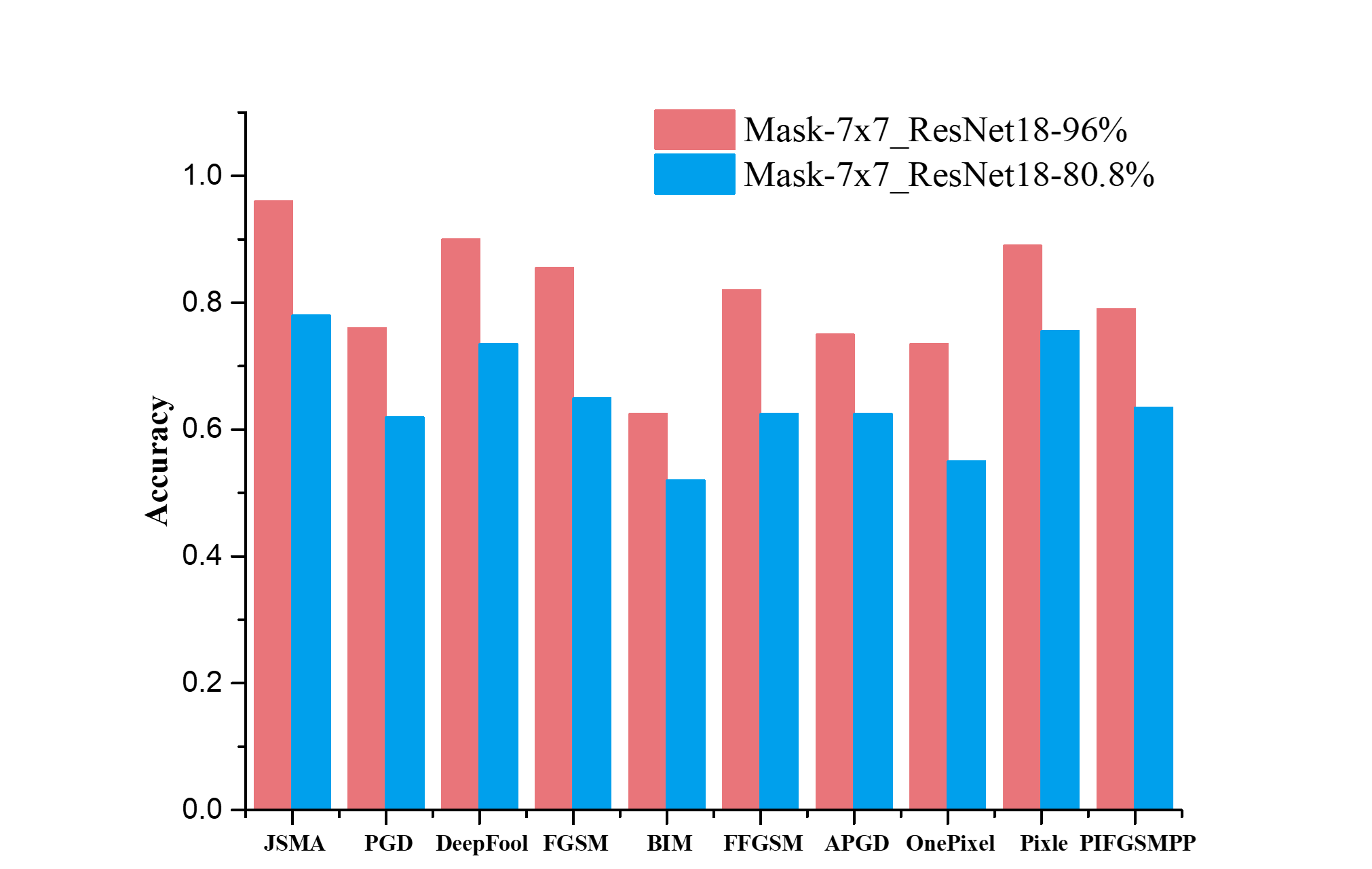}
	\captionsetup{justification=justified}
	\caption{Detection accuracy of the SWM-AED algorithm across models with different classification accuracies under various adversarial attack methods.}
	
	\label{fig:view11}
\end{figure}

\noindent\textbf{The Impact of Deep Neural Network Accuracy on the SWM-ADE Detection Algorithm.}
Table \ref{tab:attack_metrics4} presents the detection results of the SWM-AED algorithm applied to the ResNet-18 model, which achieves a training accuracy of 80.8\%. A comparative analysis with the results from Table \ref{tab:attack_metrics2} (used as a control variable) is conducted, leading to the construction of Figure~\ref{fig:view11}. This figure visually demonstrates a robust positive correlation between the model's classification accuracy and the detection performance of the SWM-AED algorithm. Specifically, as the model’s classification accuracy increases, the algorithm demonstrates improved effectiveness in detecting adversarial examples generated by different attack algorithms.  

This phenomenon can be explained from an algorithmic perspective: the SWM-AED detection mechanism relies on the stability of image features, which is quantified using the confidence scores output by the model. When a model with lower classification accuracy is used, even clean samples tend to receive lower confidence scores, leading to higher corresponding confidence entropy values. As a result, the confidence entropy distributions of adversarial and clean samples exhibit substantial overlap, making it difficult for the SWM-AED algorithm to establish an optimal decision threshold for effective differentiation. In contrast, high-accuracy models assign higher confidence scores and lower confidence entropy values to clean samples, creating a more pronounced distinction from the entropy distribution of adversarial examples. This distinction significantly enhances the detection performance of the SWM-AED algorithm by facilitating more reliable threshold selection.


\begin{figure}
	
	\centering
	\includegraphics[width=1\linewidth]{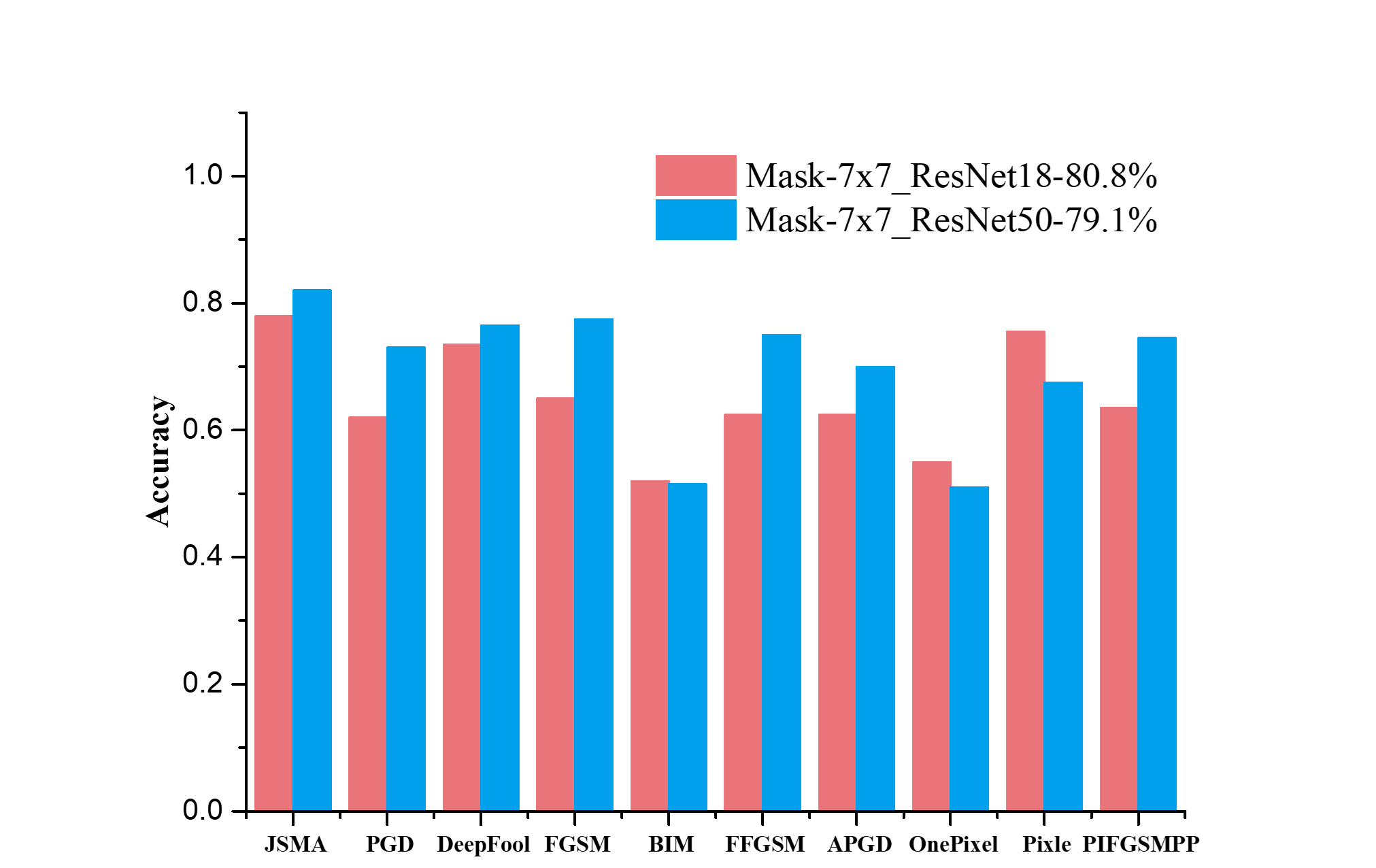}
	\captionsetup{justification=justified}
	\caption{Detection accuracy of the SWM-AED algorithm using different model layer architectures under various adversarial attack methods.}
	
	\label{fig:view12}
\end{figure}

\noindent\textbf{The Impact of Deep Neural Network Layer Architecture on the SWM-ADE Detection Algorithm.}
Tables \ref{tab:attack_metrics5} and \ref{tab:attack_metrics4} present a controlled experimental analysis designed to evaluate the impact of model depth on the performance of the proposed SWM-AED algorithm. The experimental setup of Table \ref{tab:attack_metrics5} remains largely consistent with that of Table \ref{tab:attack_metrics4}, with the sole difference being an increase in the depth of the ResNet architecture from 18 to 50 layers. The results from Tables \ref{tab:attack_metrics5} and \ref{tab:attack_metrics4} are integrated in Figure~\ref{fig:view12}, which illustrates that a deeper network architecture significantly enhances the detection performance of the SWM-AED algorithm.

This improvement can be attributed to two key factors: the algorithmic principles underlying SWM-ADE and the intrinsic characteristics of deep neural networks. First, the SWM-ADE detection mechanism is highly dependent on the model’s classification performance, which benefits from increased network depth. Second, deeper networks exhibit superior feature extraction capabilities, allowing them to capture more fine-grained representations and construct denser, more discriminative feature distributions. This leads to greater stability and reliability in classification outcomes. As a result, utilizing deeper network models for computing Sliding Mask Confidence Entropy significantly improves the effectiveness of the SWM-ADE algorithm in distinguishing between adversarial and clean samples.


\begin{figure}
	
	\centering
	\includegraphics[width=1\linewidth]{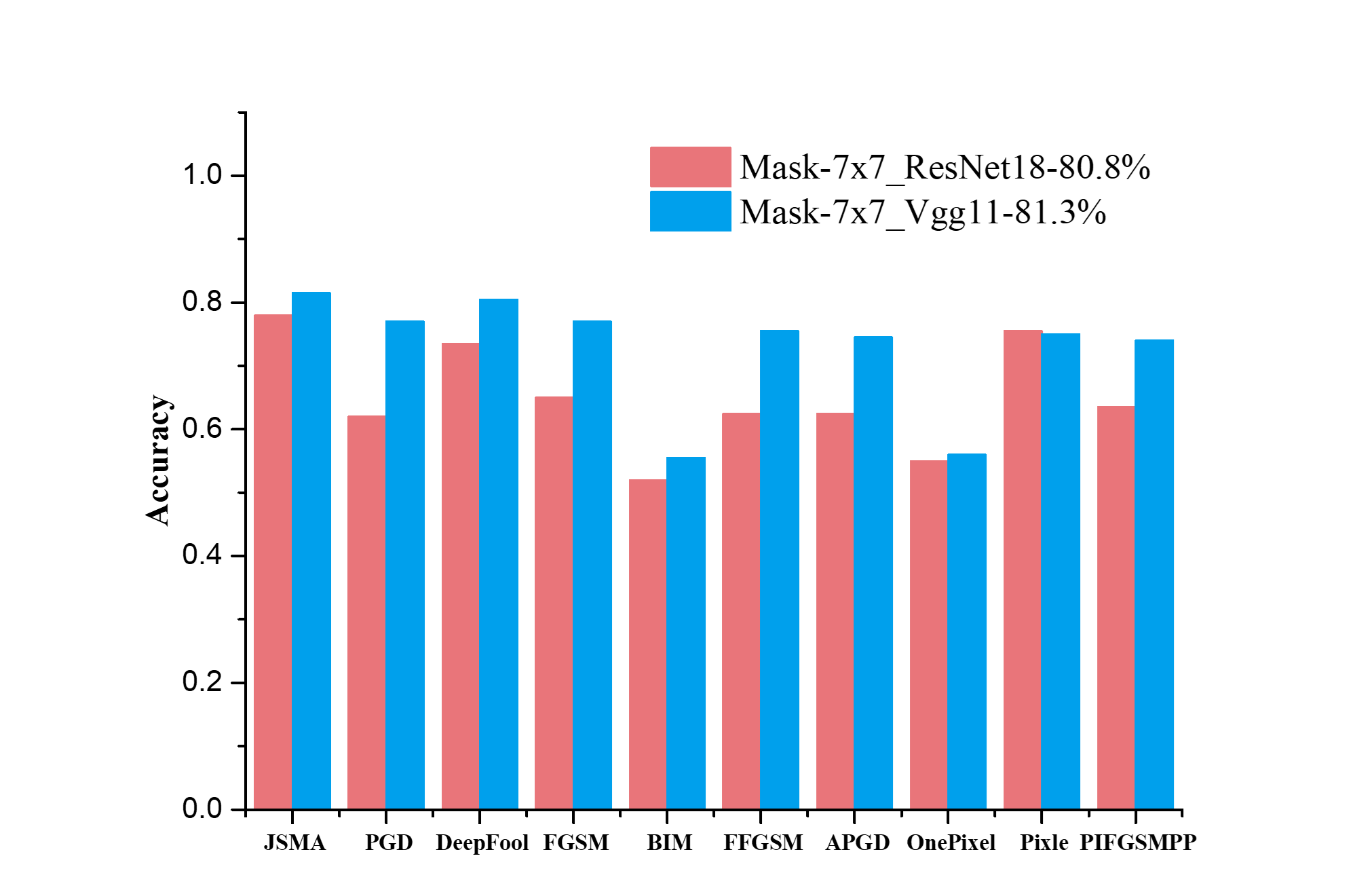}
	\captionsetup{justification=justified}
	\caption{Detection accuracy of the SWM-AED algorithm using different models under various adversarial attack methods.}
	
	\label{fig:view13}
\end{figure}

\noindent\textbf{The Impact of Different Neural Networks on the SWM-ADE Detection Algorithm.}
To further assess the detection capability of the SWM-AED algorithm across different model architectures, the experimental scope was expanded to include the VGG network, with results summarized in Table \ref{tab:attack_metrics6}. By leveraging Tables \ref{tab:attack_metrics4} and \ref{tab:attack_metrics6} as the experimental group, while controlling variables such as mask size and model accuracy, Figure~\ref{fig:view13} was generated. The experimental findings reveal that the SWM-AED algorithm demonstrates robust detection performance when the Sliding Mask Confidence Entropy (SMCE) is computed using the VGG model. This underscores the critical role of selecting high-performance deep neural networks as the foundation for SWM-AED detection.

With the continuous advancements in deep learning technology, the adoption of more powerful architectures can significantly enhance the SWM-AED algorithm’s capacity to detect adversarial examples, reflecting its potential for self-improvement alongside technological progress. By continuously updating advanced models for SMCE computation, the performance of the SWM-AED algorithm can be incrementally refined, ensuring broad applicability across diverse scenarios. Notably, the effectiveness of the SWM-AED algorithm primarily stems from its core logic—the inherent vulnerability of adversarial examples—rather than being dependent on the unique characteristics of a specific model architecture. This highlights its remarkable generalizability and suitability for widespread deployment in practical environments.

\subsubsection{Experimental Comparison with Classical Adversarial Defense Algorithms}
\begin{figure}
	\centering
	\includegraphics[width=0.75\linewidth]{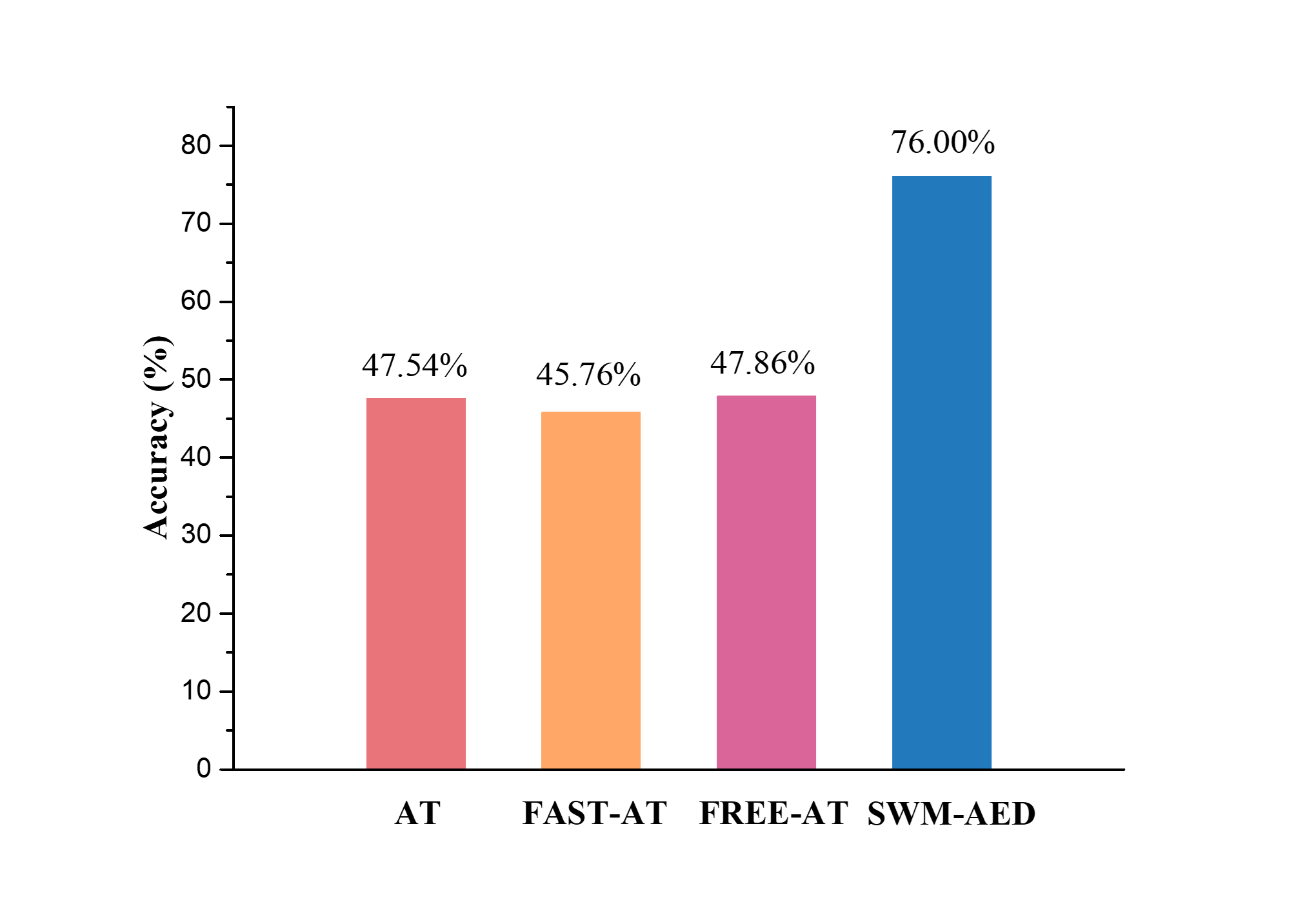}
	\caption{The figure presents a performance comparison between the SWM-AED algorithm and existing adversarial defense methods, including AT \citep{mkadry2017towards}, FAST-AT \citep{wong2020fast}, and FREE-AT \citep{NEURIPS2019_7503cfac}. It evaluates the defense success rates of these four methods against adversarial examples generated by PGD attacks on the ResNet-18 network architecture.}
	\label{fig:view14}
\end{figure}

In this section, we perform a comparative analysis between the SWM-AED algorithm and classical adversarial defense methods, namely Adversarial Training (AT) and its variants, FAST-AT and FREE-AT. As shown in Figure~\ref{fig:view14}, the experiments utilize the benchmark PGD adversarial attack algorithm and are conducted on the ResNet-18 classifier. The AT and its variants aim to defend against adversarial examples by retraining a robust ResNet-18 model, achieving a resistance rate of approximately 47\% against adversarial examples generated by the PGD algorithm. In contrast, the SWM-AED algorithm leverages SMCE computation within a deep learning model, substantially improving the defense capability to 76\%. These findings demonstrate that the SWM-AED algorithm achieves superior accuracy in detecting and defending against adversarial examples.

\section{Discussion and Conclusion}\label{sec4}

This study presents, for the first time, a systematic characterization of the intrinsic vulnerabilities of adversarial examples. While adversarial attacks are typically understood as perturbation-based manipulations that induce misclassification, we demonstrate that such perturbations inherently disrupt the local stability of input images. Specifically, the added noise perturbs the pixel value distribution, resulting in fuzzier classification boundaries. Compared to clean inputs, adversarial examples exhibit reduced semantic coherence, making classifiers more prone to prediction errors.

To quantify this instability, we introduce a novel metric: Sliding Mask Confidence Entropy (SMCE), which evaluates the sensitivity of model predictions to localized occlusions. SMCE is computed by applying a sliding window across the image and aggregating the resulting confidence entropy values. This metric captures not only the structural consistency of the input image but also the robustness of the classifier to spatially localized perturbations. Building on this insight, we propose a detection framework termed Sliding Window Mask–Adversarial Example Detection (SWM-AED), which identifies adversarial examples by assessing anomalies in their SMCE values.

The SMCE metric and SWM-AED algorithm exemplify a broader principle of deriving essential properties from observable phenomena. By leveraging localized fluctuations in model confidence, our approach reveals distinctive entropy patterns that differentiate adversarial examples from clean inputs. This capability enables robust adversarial detection without compromising classification accuracy, offering a pathway towards more secure and trustworthy AI systems. SWM-AED offers several significant innovations:
\begin{itemize}
	\item \textbf{High Practical Applicability}: SWM-AED can be seamlessly integrated into a wide range of existing deep neural network models without requiring architectural modifications. This makes it a lightweight and flexible defense mechanism that is readily deployable in real-world applications.
	
	\item \textbf{Model-Aware Adaptability}: The effectiveness of SWM-AED naturally improves with more advanced classifiers. As shown in our experiments, the SMCE metric—proposed for the first time in this work—becomes more discriminative when computed on stronger models, allowing the detection performance to scale with model robustness and accuracy.
	
	\item \textbf{Flexible and Future-Proof Design}: Unlike traditional detection methods that rely on static heuristics, SWM-AED leverages a dynamic detection mechanism grounded in a fundamental vulnerability of adversarial examples. This makes it inherently adaptable to future advances in model architecture and applicable across various domains involving deep learning.
\end{itemize}

Theoretically, this study is the first to reveal and rigorously demonstrate that adversarial examples exhibit a heightened sensitivity to occlusion compared to their corresponding benign counterparts. This finding addresses a previously overlooked characteristic of adversarial examples and contributes new insights into their intrinsic weaknesses. We believe this insight opens up new avenues for understanding the underlying mechanisms of adversarial vulnerability in deep neural networks (DNNs), potentially aiding in the development of more robust AI systems.

Notably, as demonstrated in this experiments, the detection performance of SMCE is closely tied to the robustness of the underlying classifier, making it a flexible and adaptive solution. This enables SWM-AED to be easily integrated into existing DNN-based systems across a wide range of domains. In contrast to traditional detection techniques, which often rely on fixed statistical patterns or input transformations, our method dynamically benefits from advancements in model architecture—yielding increasingly discriminative detection signals as classifier quality improves. This adaptability underscores a crucial aspect often overlooked in current AI development: the overemphasis on accuracy at the cost of systemic security. True AI reliability requires a balanced approach that prioritizes both predictive performance and robustness. The SWM-AED algorithm effectively achieves this balance, enhancing both model robustness and security while maintaining high accuracy.

\section*{Data availability}
All datasets used in this paper were obtained from public data sources and repositories. A complete list of the public data sources with links is available via GitHub at \url{https://github.com/dawei7777/SWM-AED/blob/master/README.md}.

\section*{Code availability}
The Python code used to compare the SMCE values of adversarial examples and clean samples, visualize the vulnerability of adversarial examples, and evaluate the detection accuracy of adversarial inputs using the SWM-AED algorithm is publicly available on GitHub at \url{https://github.com/dawei7777/SWM-AED}, along with documentation, DOIs, and citations \citep{li_2025_15508744}. The code is released under the MIT license without restriction.

\newpage
\appendix
%
%
%
%

\section{Proof of Property 3.2}
\renewcommand{\theequation}{A.\arabic{equation}} 
\setcounter{equation}{0} 
Property 3.2. maximum value of SMCE

\begin{align}
	H_{\text{SMCE}}(I) \leq \log_2 m \label{thm:theorem2}
\end{align}
\begin{proof}[Proof]
	In order to obtain the maximum  value of $H_{\text{SMCE}}(I)$, We formulate the following constrained  optimization problem:

	\begin{align}
		& \max_{p_{ij}} \frac{1}{n} \sum_{i=1}^{n} \left( -\sum_{j=1}^{m} p_{ij} \log_2(p_{ij}) \right) \\
		& \text{s.t.} \sum_{j=1}^{m} p_{ij} = 1 \quad \text{for $\forall $} i
	\end{align}
	Let us introduce Lagrange multipliers \( \lambda_i \) and construct the Lagrangian function as follows:
	\[
	L(p_{ij}, \lambda_i) = \frac{1}{n} \sum_{i=1}^{n} \left( -\sum_{j=1}^{m} p_{ij} \log_2(p_{ij}) \right) + \frac{1}{n} \sum_{i=1}^{n} \lambda_i \left( \sum_{j=1}^{m} p_{ij} - 1 \right)
	\]
	Solve the equations:
	\[
	\left\{
	\begin{aligned}
		\frac{\partial L(p_{ij}, \lambda_i)}{\partial p_{ij}} = 0 \\
		\sum_{j=1}^{m} p_{ij} = 1 \quad \text{for $\forall $} \quad i
	\end{aligned}
	\right.
	\]
	Obtain:
	
	\begin{align*}
		\frac{\partial L(p_{ij}, \lambda_i)}{\partial p_{ij}} 
		&= \frac{\partial \left[ -p_{ij} \log_2(p_{ij}) + \lambda_i p_{ij} \right]}{\partial p_{ij}} \\
		&= -\log_2(p_{ij}) - p_{ij} \cdot \frac{1}{p_{ij} \ln 2} + \lambda_i \\
		&= -\log_2(p_{ij}) - \frac{1}{\ln 2} + \lambda_i \\
		&= 0
	\end{align*}
	
	So \[\log_2(p_{ij}) =\lambda_i - \frac{1}{\ln 2}\]
	
	Then \begin{align} p_{ij} = 2^{\lambda_i - \frac{1}{\ln 2}} \label{eq:eq8} \end{align}

	Substituting \( p_{ij} = 2^{\lambda_i - \frac{1}{\ln 2}} \) into \( \sum_{j=1}^{m} p_{ij} = 1 \) gives:\[
	\sum_{j=1}^{m} 2^{\lambda_i - \frac{1}{\ln 2}} = m \cdot 2^{\lambda_i - \frac{1}{\ln 2}} = 1 
	\]
	
	So \begin{align} 2^{\lambda_i - \frac{1}{\ln 2}} = \frac{1}{m} \label{eq:eq9} \end{align}

	By eq. \eqref{eq:eq8} and \eqref{eq:eq9}, \begin{align} p_{ij} = \frac{1}{m} \end{align}
	
	Substituting \( p_{ij} = \frac{1}{m} \) into \( H_{\text{SMCE}}(I) \) yields:
	
	\[
	\max\  H_{\text{SMCE}}(I) = \frac{1}{n} \sum_{i=1}^{n} \left( -\sum_{j=1}^{m} \frac{1}{m} \log_2 \frac{1}{m} \right) = \frac{1}{n} \sum_{i=1}^{n} \log_2 m = \log_2 m
	\]
	
	Therefore: 
	
	\[
	H_{\text{SMCE}}(I) \leq \log_2 m
	\]
	
\end{proof}

\newpage

%

\appendix
%

\bibliographystyle{cas-model2-names}

\bibliography{sn-bibliography}

\end{document}